\RequirePackage[l2tabu, orthodox]{nag}
\documentclass[11pt]{article}

\usepackage[T1]{fontenc}

\usepackage{soul} 

\usepackage{garamond}  
\renewcommand{\rmdefault}{ugm}

\usepackage[bitstream-charter]{mathdesign}
\usepackage{amsmath}
\usepackage[scaled=0.92]{PTSans}

\usepackage[
  paper  = letterpaper,
  left   = 1.65in,
  right  = 1.65in,
  top    = 1.0in,
  bottom = 1.0in,
  ]{geometry}

\usepackage[usenames,dvipsnames]{xcolor}
\definecolor{shadecolor}{gray}{0.9}

\usepackage[final,expansion=alltext]{microtype}
\usepackage[english]{babel}
\usepackage[parfill]{parskip}
\usepackage{afterpage}
\usepackage{framed}

{\endMakeFramed}

\DeclareRobustCommand{\parhead}[1]{\textbf{#1}~}

\usepackage{lineno}

\usepackage{ragged2e}

\newcounter{parcount}

\usepackage{graphicx}
\usepackage[labelfont=bf]{caption}

\usepackage{booktabs}
\usepackage{multirow}

\usepackage[algoruled]{algorithm2e}
\usepackage{listings}
\usepackage{fancyvrb}
\fvset{fontsize=\normalsize}

\usepackage{natbib}

\usepackage[acronym,nowarn]{glossaries}
\makeglossaries

\definecolor{tangerine}{rgb}{0.95, 0.52, 0.0}
\definecolor{palebrown}{rgb}{0.6, 0.46, 0.33}
\definecolor{peru}{rgb}{0.8, 0.52, 0.25}
\usepackage[colorlinks,linktoc=all]{hyperref}
\usepackage[all]{hypcap}
\hypersetup{citecolor=peru}
\hypersetup{linkcolor=peru}
\hypersetup{urlcolor=peru}

\usepackage[nameinlink]{cleveref}
\creflabelformat{equation}{#1#2#3}
\crefname{equation}{eq.}{eqs.}  
\Crefname{equation}{Eq.}{Eqs.}

\lstdefinestyle{mystyle}{
    commentstyle=\color{OliveGreen},
    keywordstyle=\color{BurntOrange},
    numberstyle=\tiny\color{black!60},
    stringstyle=\color{MidnightBlue},
    basicstyle=\ttfamily,
    breakatwhitespace=false,
    breaklines=true,
    captionpos=b,
    keepspaces=true,
    numbers=left,
    numbersep=5pt,
    showspaces=false,
    showstringspaces=false,
    showtabs=false,
    tabsize=2
}
\usepackage[colorinlistoftodos,
           prependcaption,
           textsize=small,
           backgroundcolor=yellow,
           linecolor=lightgray,
           bordercolor=lightgray]{todonotes}

\DeclareRobustCommand{\parhead}[1]{\textbf{#1}~}

\usepackage{amsthm}  

\usepackage{bm}

\usepackage[colorinlistoftodos,
           prependcaption,
           textsize=small,
           backgroundcolor=yellow,
           linecolor=lightgray,
           bordercolor=lightgray]{todonotes}

\usepackage{graphicx}
\usepackage{caption}
\usepackage{subcaption}
\usepackage{wrapfig}

\usepackage{booktabs}
\usepackage{arydshln} 
\usepackage{multirow}
\usepackage{nicematrix}

\usepackage{listings}
\usepackage{fancyvrb}
\fvset{fontsize=\normalsize}

\usepackage[nameinlink]{cleveref}
\creflabelformat{equation}{#1#2#3}
\crefname{equation}{eq.}{eqs.}  
\Crefname{equation}{Eq.}{Eqs.}

\usepackage{listings}
\lstdefinestyle{alp_style}{
    commentstyle=\color{OliveGreen},
    numberstyle=\tiny\color{black!60},
    stringstyle=\color{BrickRed},
    basicstyle=\ttfamily\scriptsize,
    breakatwhitespace=false,
    breaklines=true,
    captionpos=b,
    keepspaces=true,
    numbers=none,
    numbersep=5pt,
    showspaces=false,
    showstringspaces=false,
    showtabs=false,
    tabsize=2
}

\usepackage{capt-of}

\usepackage{lipsum}

\theoremstyle{remark}
\newtheorem*{lemma*}{Lemma}

\DeclareMathOperator*{\argmax}{arg\,max}

\newcommand{\bp}{\bm{p}}
\newcommand{\bX}{\bm{X}}
\newcommand{\bK}{\bm{K}}
\newcommand{\bC}{\bm{C}}

\newcommand{\bx}{\bm{x}}

\usepackage{authblk}
\usepackage{enumitem} 
\usepackage{algorithmic}
\usepackage{tabularx}
\usepackage{booktabs}
\usepackage{pifont}
\usepackage[T1]{fontenc}
\usepackage{multirow}
\usepackage{threeparttable}

\title{\textbf{The Vendiscope:\\ An Algorithmic Microscope For Data Collections}}

\author[1, 3]{Amey P. Pasarkar}
\author[2, 3]{Adji Bousso Dieng}
\affil[1]{Lewis-Sigler Institute For Integrative Genomics, Princeton University}
\affil[2]{Department of Computer Science, Princeton University}
\affil[3]{\href{https://vertaix.princeton.edu/}{Vertaix}}

\begin{document}
\maketitle

\begin{abstract}
\noindent The evolution of microscopy, beginning with its invention in the late 16th century, has continuously enhanced our ability to explore and understand the microscopic world, enabling increasingly detailed observations of structures and phenomena. In parallel, the rise of data-driven science has underscored the need for sophisticated methods to explore and understand the composition of complex data collections. This paper introduces the \emph{Vendiscope}, the first \emph{algorithmic microscope} designed to extend traditional microscopy to computational analysis. The Vendiscope leverages the Vendi scores—a family of differentiable diversity metrics rooted in ecology and quantum mechanics—and assigns weights to data points based on their contribution to the overall diversity of the collection. These weights enable high-resolution data analysis at scale. We demonstrate this across biology, materials science, and machine learning (ML). We analyzed the 250 million protein sequences in the protein universe, discovering that over 200 million are near-duplicates and that AlphaFold fails on proteins with Gene Ontology (GO) functions that contribute most to diversity. Applying the Vendiscope to the Materials Project database led to similar findings: more than $85\%$ of the crystals with formation energy data are near-duplicates and ML models perform poorly on materials that enhance diversity. Additionally, the Vendiscope can be used to study phenomena such as memorization in generative models. We used the Vendiscope to identify memorized training samples from 13 different generative models spanning several model classes and found that the best-performing generative models often memorize the training samples that contribute least to diversity. Our findings demonstrate that the Vendiscope can serve as a powerful tool for data-driven science, providing a systematic and scalable way to identify duplicates and outliers, as well as pinpointing samples prone to memorization and those that models may struggle to predict—even before training.\\


\end{abstract}

\section*{Main}
\label{sec:main}

In recent years, many scientific fields have transitioned to data-driven paradigms, where the sheer scale and complexity of datasets often outpace traditional analytical techniques. Biology, materials science, and artificial intelligence (AI) are prime examples of disciplines now dominated by vast, high-dimensional data collections that are critical for advancing knowledge. However, these datasets are frequently rife with redundancies, biases, and oddities that traditional methods may struggle to capture or analyze at scale ~\citep{ding2024protein, elazar2023s, griffiths2021dataset, hart2024trust, liu2021visually, vickers2021quantifying}. Furthermore, in the rapidly evolving field of AI, there is an increasing call for data-centric methodologies that focus on improving data quality rather than solely refining models~\citep{jones2024rethinking, oala2023dmlr}. As AI systems become more deeply integrated into scientific and industrial applications, the need for computational tools capable of scrutinizing the underlying composition of datasets has never been more pressing \citep{longpre2024large, sambasivan2021everyone}. 

This paper introduces the concept of \emph{algorithmic microscopes}, computational tools that allow scientists to "zoom in" on datasets to make discoveries. What should an algorithmic microscope do? At its core, it should help uncover new insights by carefully examining the dataset itself, without the need to build a model of the data. Just as a traditional microscope reveals unseen details in the physical world, an algorithmic microscope should reveal the hidden details of a data collection, enabling the discovery of rare items, outliers, and unexpected patterns. These elements often hold the key to breakthroughs or findings that warrant deeper investigation. An equally important aspect of data-driven discovery is identifying redundancies in data. Redundant data points can provide insight into the underlying structure of the dataset, e.g. in the form of clusters. By detecting where redundancies occur, an algorithmic microscope may identify biased data sources and reveal areas where the data collection process may need to be expanded. Another critical approach in the discovery process is the comparison of data collections. By jointly analyzing datasets, an algorithmic microscope ought to be able to reveal commonalities or gaps that would be invisible when the datasets are examined in isolation. For example, comparing a training set with the outputs of a generative model can uncover memorized samples, revealing important aspects of model behavior and generalization. Finally, an algorithmic microscope should be able to rank data points by their rarity or commonality. This ranking can serve as a guiding principle for efficient data processing, model training, and discovery. 

In this paper, we demonstrate that all of these capabilities---detecting outliers, identifying redundancies, comparing data collections, and ranking by rarity---can be achieved at scale by answering a single question: What is the contribution of each data point to the overall diversity of the collection? We introduce the Vendiscope, the first algorithmic microscope designed to answer this question. The Vendiscope maximizes the probability-weighted Vendi Score (pVS) of an input collection. The pVS, introduced in \cite{friedman2023vendi}, is a similarity-based diversity metric that measures the diversity of a collection of elements sampled according to a discrete probability distribution, where each element is assigned a probability. In this work, the Vendiscope learns these probabilities, which correspond to the contributions of each data point to the overall diversity of the collection. 

To scale the Vendiscope to large data collections, such as the protein universe, we employ projective gradients, leverage pre-trained feature extractors to compute cosine similarity for the Vendi Score calculation, and use parallel computing. These scaling strategies enable the Vendiscope to operate with linear complexity in both space and time, making it feasible to analyze extremely large datasets.

We applied the Vendiscope to three different domains: biology, materials science, and AI. We first analyze the universe of protein sequences, which consists of nearly $250$ million entries. Despite the scale of this dataset, the Vendiscope can provide an end-to-end analysis within 2 hours on a single compute node comprised of $8$ NVIDIA A6000 GPUs. We find that AlphaFold struggles to provide accurate structural predictions for sequences that contribute most to diversity. We highlight some of the genes and functions enriched among these sequences. Additionally, we find that over $80\%$ of the entire dataset are near-duplicates. This suggests that while the universe of protein sequences is vast, its diversity is significantly lower. 

A similar pattern emerges on the Materials Project (MP) database, a benchmark dataset containing almost $170,000$ material structures and various properties. We show how three popular models---ALIGNN \citep{choudhary2021atomistic}, CGCNN \citep{xie2018crystal}, and DeeperGATGNN \citep{omee2022scalable}---all fail in the same manner: their formation energy and band-gap property prediction errors are significantly higher on materials that contribute most to diversity. Furthermore, we expect modeling efforts to improve as MP continues to grow, however, we highlight the need for the addition of unique, non-redundant materials: over 85\% of crystals with formation energy data are near-duplicates in the MP database. 

We also show how the Vendiscope can reveal patterns of memorization across $13$ state-of-the-art image generative models trained on CIFAR10 and spanning popular model classes such as diffusion models, GANs, VAEs, and flows. The Vendiscope reveals that the models with the best-looking outputs---those that achieve high human error rate---tend to memorize data points that contribute least to the diversity of the training set. 

Our results reveal systemic challenges in modeling data points that contribute most to diversity, across domains. This is exacerbated by a staggering amount of data redundancy in these domains. The Vendiscope can serve as a powerful tool for data-driven discovery and help create more balanced, high-quality datasets in an era of data-driven science.

\section*{The Vendiscope}
\label{sec:vendiscope}

Consider a collection of $N$ elements $(\bx_1, \dots, \bx_N)$. Let $k(\cdot, \cdot)$ denote a positive semi-definite kernel that measures the similarity between any two elements, and such that $k(\bx_i, \bx_i) = 1$ $\forall i$. Denote by $\bK$ the similarity matrix induced by the kernel $k(\cdot, \cdot)$. Its element at row $i$ and column $j$ is $K_{ij} = k(\bx_i, \bx_j)$. Since $k(\cdot, \cdot)$ is positive semi-definite, $\bK$ is positive semi-definite and has nonnegative eigenvalues which we denote by $\lambda_1, \dots, \lambda_N$. Let $\bp = (p_1, \dots, p_N)$  denote a discrete probability distribution over the collection $(\bx_1, \dots, \bx_N)$. Define $\tilde\bK_p = \text{diag}(\sqrt{\bp}) \bK \text{diag}(\sqrt{\bp})$ and let $\eta_{1p}, \dots, \eta_{Np}$ denote the eigenvalues of $\tilde\bK_p$. \cite{friedman2023vendi}
 define the probability-weighted Vendi Score (pVS) of the collection as
 \begin{align}\label{eq:pvs}
     \text{pVS}_k(\bx_1, \dots, \bx_N, \bp) &= \exp\left(-\sum_{i=1}^{N} \eta_{ip} \log \eta_{ip}\right).
 \end{align}
 This can be generalized using the \text{R\'enyi} entropy~\citep{pasarkar2024cousins},
 \begin{align}\label{eq:pvs-general}
     \text{pVS}_k(\bx_1, \dots, \bx_N, \bp) &= \exp\left(\frac{1}{1-q}\log \sum_{i \in \text{supp}(\eta)}^{} \eta_{ip}^q\right),
 \end{align}
where $\text{supp}(\eta)$ denotes the set of non-zero eigenvalues of $\tilde\bK_p$ and $q \geq 0$ is the order of the pVS. Settting $q = 1$ recovers Eq. \ref{eq:pvs}. We refer the reader to \citet{pasarkar2024cousins} for a detailed discussion of how $q$ influences the Vendi score. 

The Vendiscope considers $\bp$ as an unknown probability distribution to be learned by maximizing Eq. \ref{eq:pvs-general},
\begin{align}
    \bp^* &= \argmax_{\bp} \text{ pVS}_k(\bx_1, \dots, \bx_N, \bp) \text{ such that } \sum_{i=1}^{N} p_i = 1
    .
\end{align}
Optimizing Eq. \ref{eq:pvs-general} will assign a higher probability to the rarest samples, and a lower probability to the most common ones. Indeed, for all orders of $q$, Equation \ref{eq:pvs-general} is maximized when the eigenvalues $\eta_{1p}, \dots, \eta_{Np}$ are the same. Therefore, to successfully optimize Equation \ref{eq:pvs-general} the probabilities should be learned such that all eigenvalues are within a small $\epsilon$ distance of each other: $\eta_{\min} \leq \eta_{ip} \leq \eta_{\min}+\epsilon$, where $\epsilon>0$ and $\eta_{\min}$ denotes the minimum eigenvalue. We assume without loss of generality that $\eta_{\min}$ is non-zero. We can link the uniformity of the eigenvalues to the Vendiscope's learned probabilities using the Gershgorin Circle Theorem \citep{varga2011gervsgorin}. From this theorem, we know that the eigenvalues of  $\tilde\bK_p$ are located in discs with radii determined by the row-sums. Define $C_j = \sum_{i \ne j}^N K_{ij} \sqrt{p_i}$, which corresponds to a sum of weighted similarities between one sample and the rest of the dataset. Then, for each eigenvalue $\eta_{ip}$, there exists a row index $j \in \left\{1,\dots,N\right\}$ such that 
\begin{align}
    | \eta_{ip} - p_j | \leq \sqrt{p_j} C_j 
\end{align}
\citet{varga2011gervsgorin} additionally states in Theorem 1.6 that if a set of $L$ discs is disjoint from all other discs, it must contain $L$ eigenvalues. As a result, if there exists a single sample $\bx_j$ with disc centered at $p_j$ that is disjoint from all other discs and is not within $\sqrt{p_j} C_j$ of the eigenvalue interval $\left[\eta_{\min}, \eta_{\min}+\epsilon\right]$, it would contain an eigenvalue that violates our uniformity assumption. We therefore expect all discs to be tightly clustered around the eigenvalue interval.

In order to construct such discs, the highest probabilities $p_j$ must be assigned to the samples $\bx_j$ with the smallest weighted row-sums $C_j$. Otherwise, any disc with small $C_j$ and $p_j$ will have a small radius and be far away from the eigenvalue interval, creating a disjoint disc. Since samples with low $C_j$ are those that are most distinct from the rest of the dataset, particularly other high-probability samples, assigning high probability to them ensures the rarest samples receive the greatest weight in the optimal $\bp^*$.

Details on the Vendiscope's gradient-based optimization algorithm, its implementation, and scalability are provided in \nameref{sec:details}.

\parhead{Detecting rare elements.}
We call \emph{rare} elements those data points that contribute most to the diversity of the collection. As demonstrated earlier, these are the data points to which the Vendiscope assigns the highest probabilities. Our experiments also show that these data points tend to be the ones that models may struggle to predict. Indeed, we find that the data points that are assigned the lowest probabilities by the Vendiscope yield the best model predictions.

\parhead{Detecting duplicates.}
Duplicates in data will contribute to the diversity of a dataset almost identically. These duplicates should therefore have very similar probabilities. This insight motivates how we detect duplicates in Algorithm \ref{alg:duplicate}. Importantly, we do not need to calculate all the $N^2$ pair-wise similarities in the data and can instead focus on data points that the Vendiscope assigns similar probabilities. More specifically, we find redundant data points by only computing similarities between each data point and its $m$ closest neighbors, where closeness is measured using the assigned probabilities from the Vendiscope. Choosing $m$ large comes at a higher computational cost. We find that values of $m$ in the order of $1-2\%$ of the size of the dataset are sufficient for analyzing large-scale datasets with hundreds of millions of data points. At this scale, the Vendiscope can identify over $95\%$ of all duplicates at only a fraction of the cost of computing all pairwise similarities.

\parhead{Detecting memorization.}
A naive way to detect memorized data points from a collection of outputs from a generative model is to compare each output with all the elements in the training set. This has a huge computational cost for large training sets. Our empirical study suggests a more scalable and systematic approach for detecting memorized data points: apply the Vendiscope to the generated collection and inspect outputs that are assigned the lowest probabilities by the Vendiscope. We find across 13 generative models spanning different model classes that the outputs with the lowest probabilities have higher similarity with the training set, with similarities reaching close to 1 for generative models with higher perceptual quality. \cite{pasarkar2024cousins} found that image generative models with high human error rates---those whose outputs are harder to differentiate from real images by human judges---are those that memorize training data and produce duplicates around those points. Our findings in this paper confirm that and go further to characterize which samples are memorized: the ones that contribute least to diversity. The finding also holds when applying the Vendiscope to the training set instead of the generated collection. The training points assigned the lowest probabilities by the Vendiscope have higher similarities with outputs from the generative models. These findings have implications for discovery but also issues related to copyrights and privacy. Finally, they also suggest the Vendiscope can help detect data points that may be prone to memorization even before training.

\begin{figure}[t]
    \centering
    \includegraphics[trim={1.cm 4.cm 1.2cm 6.5cm},clip, width=0.9\textwidth]{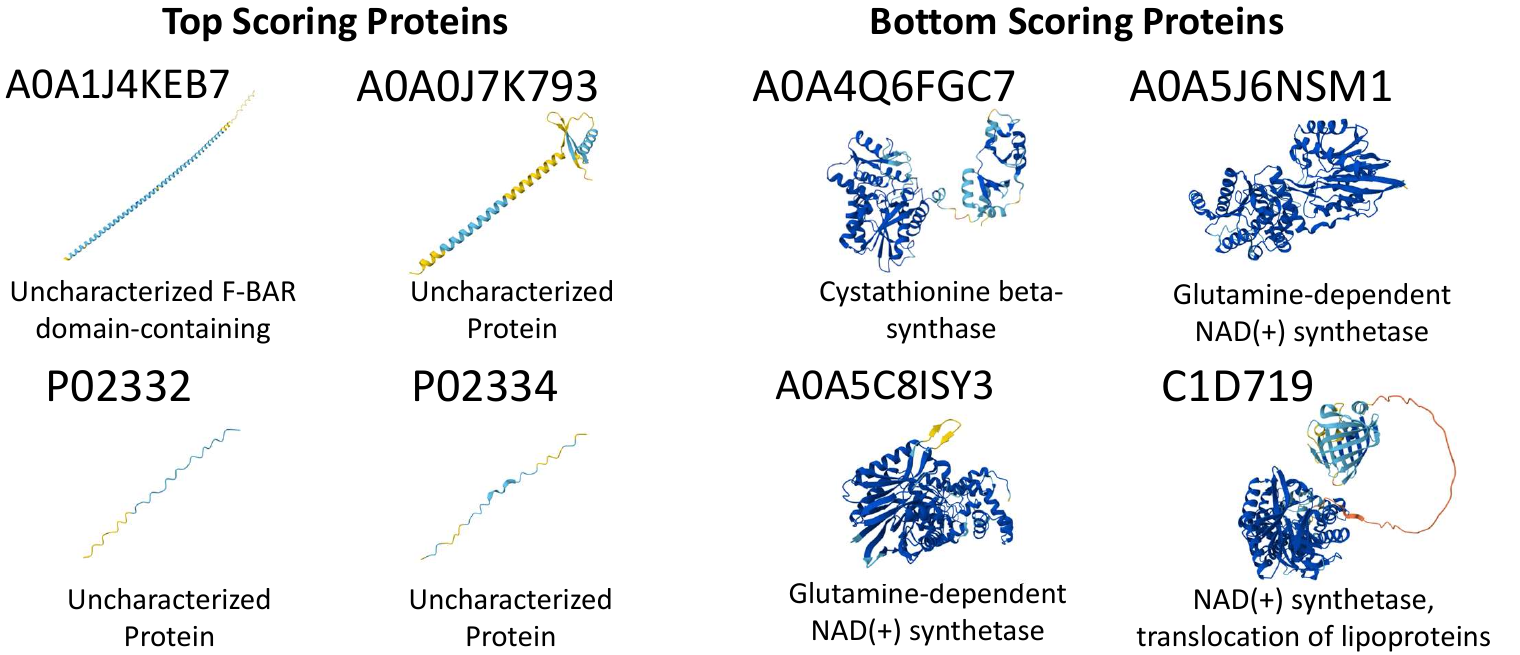}
    \hfill
    \caption{The rarest (top-scoring) proteins and the (low-scoring) proteins that contribute least to diversity, as identified by the Vendiscope, along with their corresponding AlphaFold predicted structures. Rare proteins are mostly uncharacterized or are biologically unrealistic. For example, one of the identified rare proteins misses the characteristic banana shape in the F-BAR domain. In contrast, bottom-scoring proteins are involved in fundamental pathways such as NAD(+) synthesis and transsulfuration.}
    \label{fig:topbottom4_proteins}
\end{figure}

\section*{Analyzing the Universe of Protein Sequences}
\label{sec:proteins}

The number of annotated protein sequences has increased 100x over the past two decades. Today, the UniProt database, also referred to as the \textit{protein universe}, contains $250$ million annotated sequences. ML models such as AlphaFold, ProtBert, and ProtT5 use UniProt for various modeling purposes~\citep{jumper2021highly, brandes2022proteinbert, elnaggar2021prottrans}. However, methods to analyze the composition of this database are needed. 

Currently, MMseqs2 is commonly used to cluster protein sequences based on sequence similarity and is the most popular approach for curating smaller versions of the UniProt database, such as UniRef90 and UniRef50~\citep{steinegger2018clustering}. Researchers rely on these smaller datasets for faster sequence searches, more efficient model training, and improved functional annotations~\citep{sieber2018recovery, suzek2015uniref}. However, we found that MMseqs2 produces overly conservative sequence clusters; it assigns many similar sequences to separate clusters.~This prevents further reductions in dataset size and limits computational efficiency gains. Furthermore, MMseqs2 has limitations: the algorithm only detects near-duplicate sequences and does not provide insights into which protein sequences are over- or under-represented in the dataset.

\begin{figure}[t]
    \centering
    \includegraphics[width=0.95\textwidth]{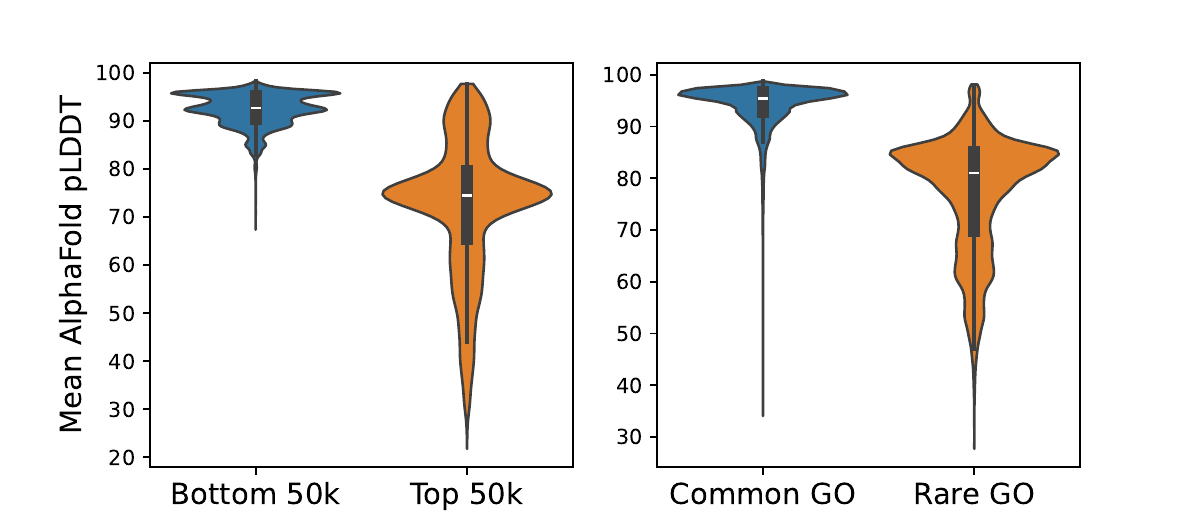}
    \caption{AlphaFold confidence is worse on rare protein sequences. \textbf{Left:} Violin plot of average pLDDT for the top (most rare) and bottom (most common) $50,000$ sequences is shown. \textbf{Right:} Violin plot of AlphaFold confidences for proteins with certain GO functions. We select $10$ GO functions that are primarily present among low-scoring proteins ('Common GO') and $10$ GO functions that are enriched among high-scoring proteins ('Rare GO'). GO functions are shown in Figure \ref{fig:TopBottomEnrichment}.
    }
    \label{fig:Alphafold_confidence}
\end{figure}

Here, we show that the Vendiscope can simultaneously produce a curated version of UniProt that is less than half the size of MMseqs2's output, while also providing deeper insights into the UniProt database by detecting and characterizing functional and evolutionary patterns among sequences. Unlike MMseqs2, the Vendiscope forms significantly larger clusters without compromising biological accuracy. Moreover, the Vendiscope sheds light on model behavior by highlighting sequences that models, such as AlphaFold, struggle to predict, making it a valuable tool for assessing model performance and guiding future modeling and data collection strategies.

We apply the Vendiscope to the entire protein universe. We generate embeddings of the protein sequences using the ProtT5-XL-UniRef50 model \citep{elnaggar2021prottrans}, averaging over all per-residue embeddings to obtain a single vector representation per protein. Using the algorithms described in \nameref{sec:details}, we are able to run the Vendiscope and detect all duplicate clusters in under 2 hours using a single compute node equipped with 8 NVIDIA A6000 GPUs. We expect even faster speeds with optimized data loading procedures. Running MMSeqs2 on the protein universe also requires 2 hours using 40 CPU cores, although we show below that the Vendiscope provides a more thorough analysis. 

\parhead{AlphaFold struggles with proteins that contribute most to diversity.} Figure~\ref{fig:topbottom4_proteins} highlights examples of the rare and common protein structures identified by the Vendiscope. The rare sequences, those that contribute most to the diversity of the database, tend to correspond to poorly annotated proteins with questionable AlphaFold predictions, whereas the 
lowest-scoring proteins, those that contribute least to the diversity of the protein universe, are those involved in the fundamental NAD+ synthesis and transsulfuration pathways and seem to have better structural predictions. We confirm the effect of rarity on prediction quality in Figure~\ref{fig:Alphafold_confidence}: the AlphaFold prediction confidence, as determined by the average predicted local distance difference test (pLDDT) over each sequence, declines significantly for the rarest sequences.

\begin{figure}[!t]
    \centering
    \includegraphics[trim={0cm 0 0cm 0},clip,width=0.95\textwidth]{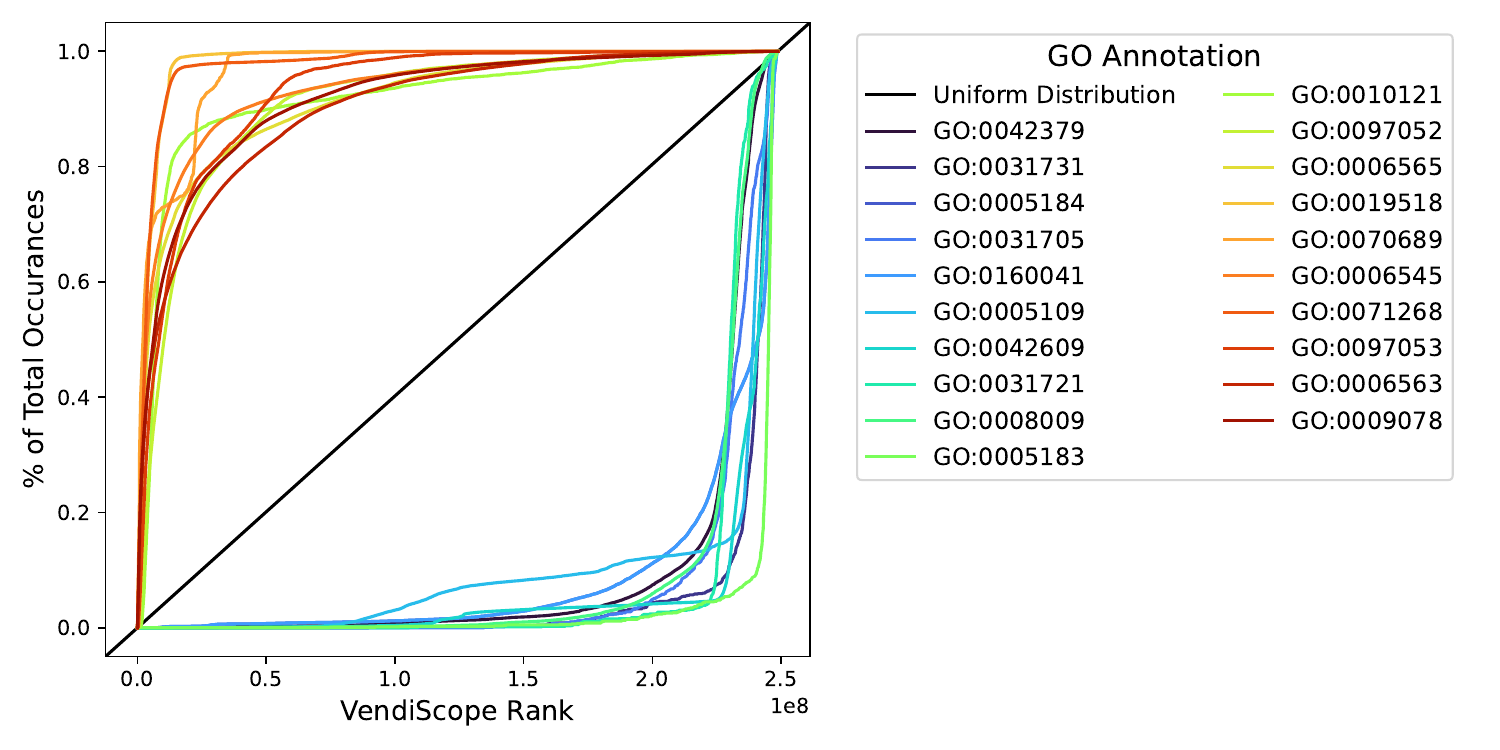}
    \hfill
    \caption{Various selected Gene Ontology (GO) functions that are enriched among highly-ranked and low-ranked proteins. All displayed functions concentrated in rare proteins have roles in protein binding (GO:0005515), whereas all displayed functions in low-ranked proteins fall under amino acid metabolic processes (GO:0006520). }
    \label{fig:TopBottomEnrichment}
\end{figure}

Structural predictions are also markedly worse for proteins with functions that are primarily found in rare sequences compared to those common in the dataset. Figure~\ref{fig:TopBottomEnrichment} shows the GO functions present in some of these rare and common protein sequences identified by the Vendiscope. We find various GO terms related to protein binding concentrated among rare sequences, whereas functions related to amino acid metabolic processes are present among common sequences. 

Importantly, the weights assigned by the Vendiscope are not merely a reflection of the prevalence of each function, but rather a result of how biological functions evolve. Amino acid metabolism, for example, relies on structurally flexible enzymes that have been repeatedly co-opted for new functions throughout evolution. This has led to a proliferation of homologous sequences across organisms and other functions, resulting in highly similar proteins that are ranked lower by the Vendiscope \citep{jensen1976enzyme}. In contrast, we find various protein binding functions that are present primarily in rare protein sequences. These include ligands for chemokine receptors (GO:0031731, GO:0042379), bombesin receptors (GO:0031705) and frizzled binding (GO:0005109). Each of these ligands bind to G protein-coupled receptors, which are known to have highly diverse ligands that vary in both size and sequence. Many genes encoding these functions also have lethal phenotypes, which impose strong evolutionary constraints \citep{wang2016frizzled, zlotnik2006chemokine}. This negative selection pressure prevents the emergence of new, highly similar sequences with different functions. As a result, proteins with these functions are diverse from one another and from the rest of the protein universe. The Vendiscope captures these evolutionary phenomena that led to the diversity of the protein universe today. 

\begin{figure}[t]
    \centering
    \includegraphics[trim={0.4cm 0 0.4cm 0.05cm}, width=\textwidth]{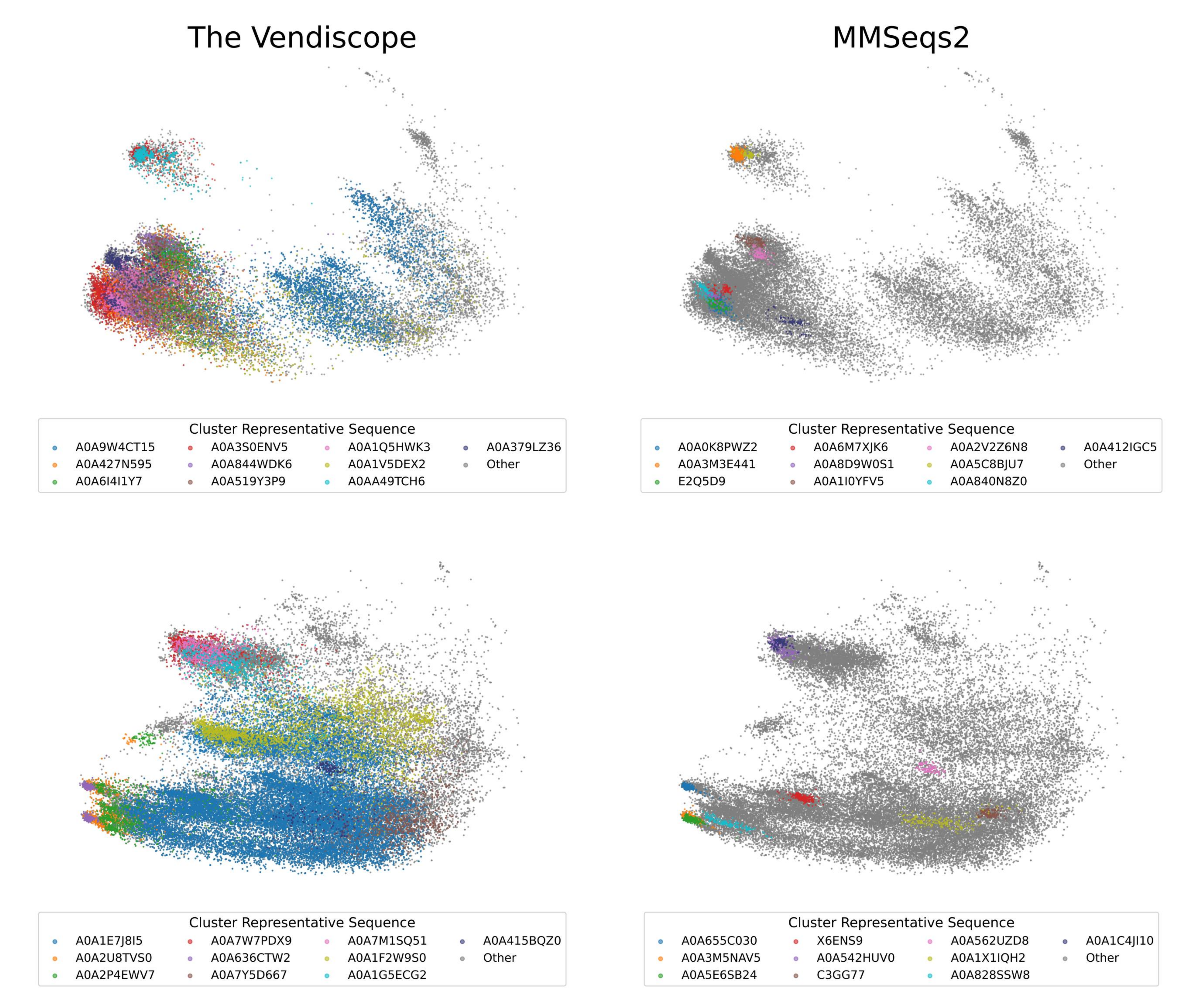}
    \caption{The Vendiscope identifies duplicates more accurately than MMseqs2, as demonstrated by the large protein clusters with consistent annotations it finds. \textbf{Top:} PCA scatter plot of all proteins originating from the ahcY gene, with duplicate clusters from the Vendiscope (left) and MMseqs2 (right). The $10$ clusters with the most proteins from the ahcY gene are shown for both methods. \textbf{Bottom:} PCA scatter plot of all proteins annotated with the GO:0003862 function (3-isopropylmalate dehydrogenase activity), with duplicate clusters from the Vendiscope (left) and MMseqs2 (right). The $10$ clusters with the most proteins containing this function are shown for both methods.}
    \label{fig:Protein_Dups}
\end{figure}

\parhead{The Vendiscope identifies redundant sequences in the protein universe.} With a search-range of $m=2,000,000$ in Algorithm~\ref{alg:duplicate}, we identify $21,003,854$ clusters containing $210,372,272$ proteins above a $s=0.9$ similarity threshold. For comparison, using the same widely adopted threshold of $0.9$, MMseqs2 identifies $29,540,400$ clusters that encompass only $127,545,233$ proteins. Figure~\ref{fig:Protein_Dups} shows how the clusters identified by the Vendiscope have clear biological interpretations and are significantly larger than those identified by MMseqs2.

To further benchmark the quality of the clusters identified by the Vendiscope, we measure the consistency of the functions of the proteins in each cluster. Each protein has a list of GO annotations that describe all of the protein's known functions \citep{ashburner2000gene, gene2023gene}. To measure the similarity between two GO terms, we record the reciprocal of the distance between GO terms on the corresponding GO tree, as described in \citep{sangar2007quantitative}. To then compute the similarity between pairs of proteins $P_1$ and $P_2$, we must compare two lists of GO terms. We use the Average-Best-Match approach by \cite{zhao2018gogo}. Suppose $P_1$ has $m$ terms go$_{11}$, go$_{12}$, \dots, go$_{1m}$ and $P_2$ has $n$ terms go$_{21}$, go$_{22}$, \dots, go$_{2n}$. The similarity between $P_1$ and $P_2$ is defined as 
\begin{align}\label{eq:prot-sim}
    k'(P_1, P_2) &= \frac{1}{m+n} \left( \sum_{i=1}^m \max_{\text{go}_{1i}} k (\text{go}_{1i}, \text{go}_{2j}) + \sum_{j=1}^n \max_{\text{go}_{2j}} k(\text{go}_{1i}, \text{go}_{2j}) \right).
\end{align}
where $k(\cdot, \cdot)$ is a similarity kernel. Eq. \ref{eq:prot-sim} provides a measure of similarity between the functional annotations of a pair of proteins. To then obtain a measure of consistency for a given cluster, we compute the average semantic similarity between a cluster's representative sequence and all other sequences in the cluster. In the Vendiscope, we define the representative sequence for each cluster as the sequence whose closest to the cluster's centroid. For context, the representative sequence in a MMseqs2 cluster is the cluster's longest sequence. We find that the clusters identified by the Vendiscope have an average semantic similarity of $0.942 \pm 0.105$, while those identified by MMseqs2 have an average of $0.985 \pm 0.049$ semantic similarity. Both similarities are quite high and likely suffer from how certain proteins may have poor annotations. Nevertheless, the Vendiscope is within one standard deviation of MMseqs2 in terms of semantic similarity while still identifying $65\%$ more proteins with near-duplicates.

\section*{Analyzing The Materials Project Database}
\label{sec:materials}

Next, we use the Vendiscope to analyze the composition of the Materials Project database (v2024.12.18). The Materials Project is the result of a significant computational effort to calculate the properties of many materials \citep{jain2013commentary}. This database has been instrumental in training ML models for materials property prediction and continues to grow. The prioritization of which materials are added has significant implications for the quality of future models. Using the Vendiscope on three popular models—ALIGNN \citep{choudhary2021atomistic}, CGCNN \citep{xie2018crystal}, and DeeperGATGNN \citep{omee2022scalable}—we characterize the materials in the Materials Project, reveal potential biases within the database, and identify patterns of model failure for property prediction. Details for model training are provided in the Appendix.  

\begin{figure}[tp]
    \centering
    \begin{minipage}[c]{0.49\linewidth}
    \includegraphics[trim={0cm 0cm 0cm 0cm},width=\textwidth]{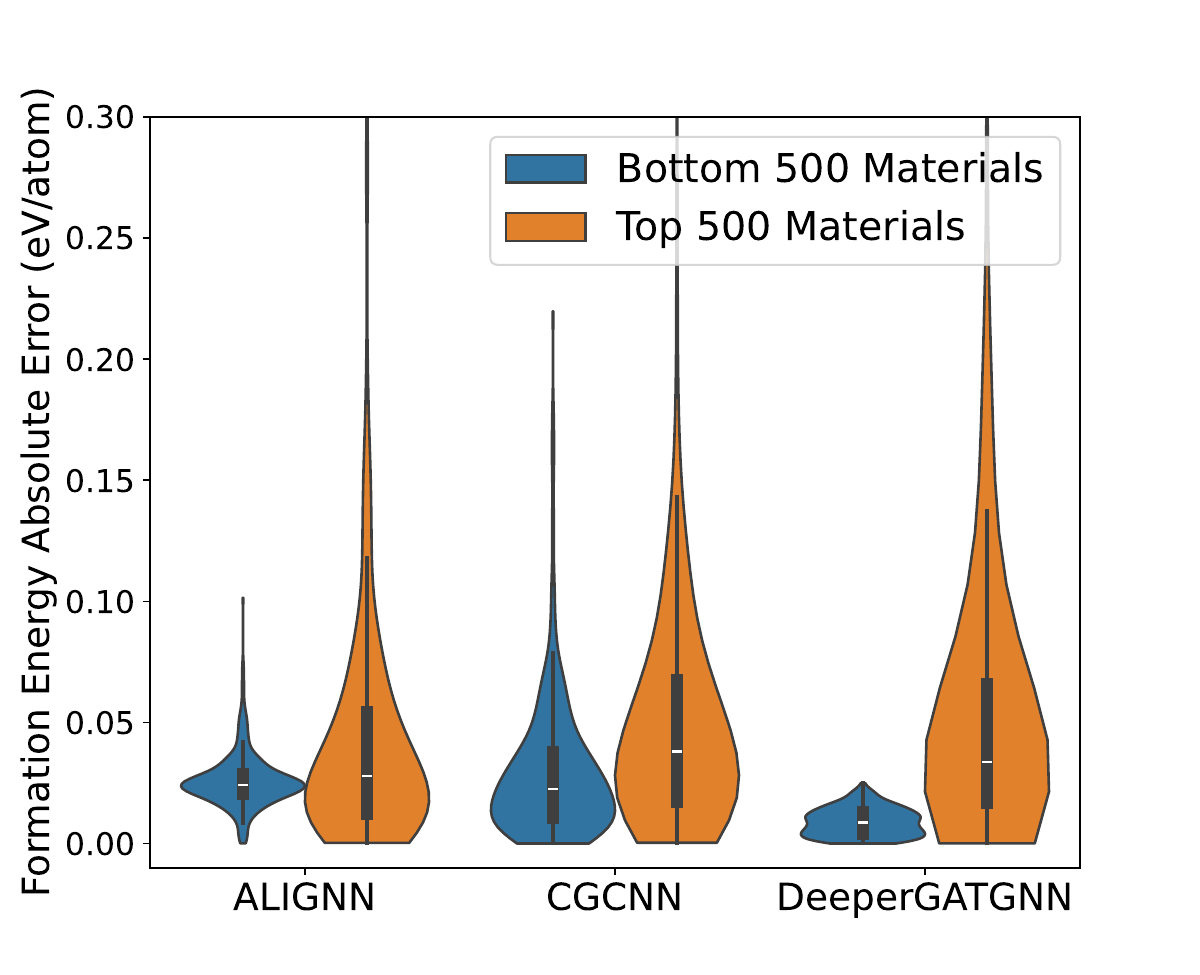}
    \end{minipage}
    \begin{minipage}[c]{0.49\linewidth}
    \includegraphics[trim={0cm 0cm 0cm 0cm},width=\textwidth]{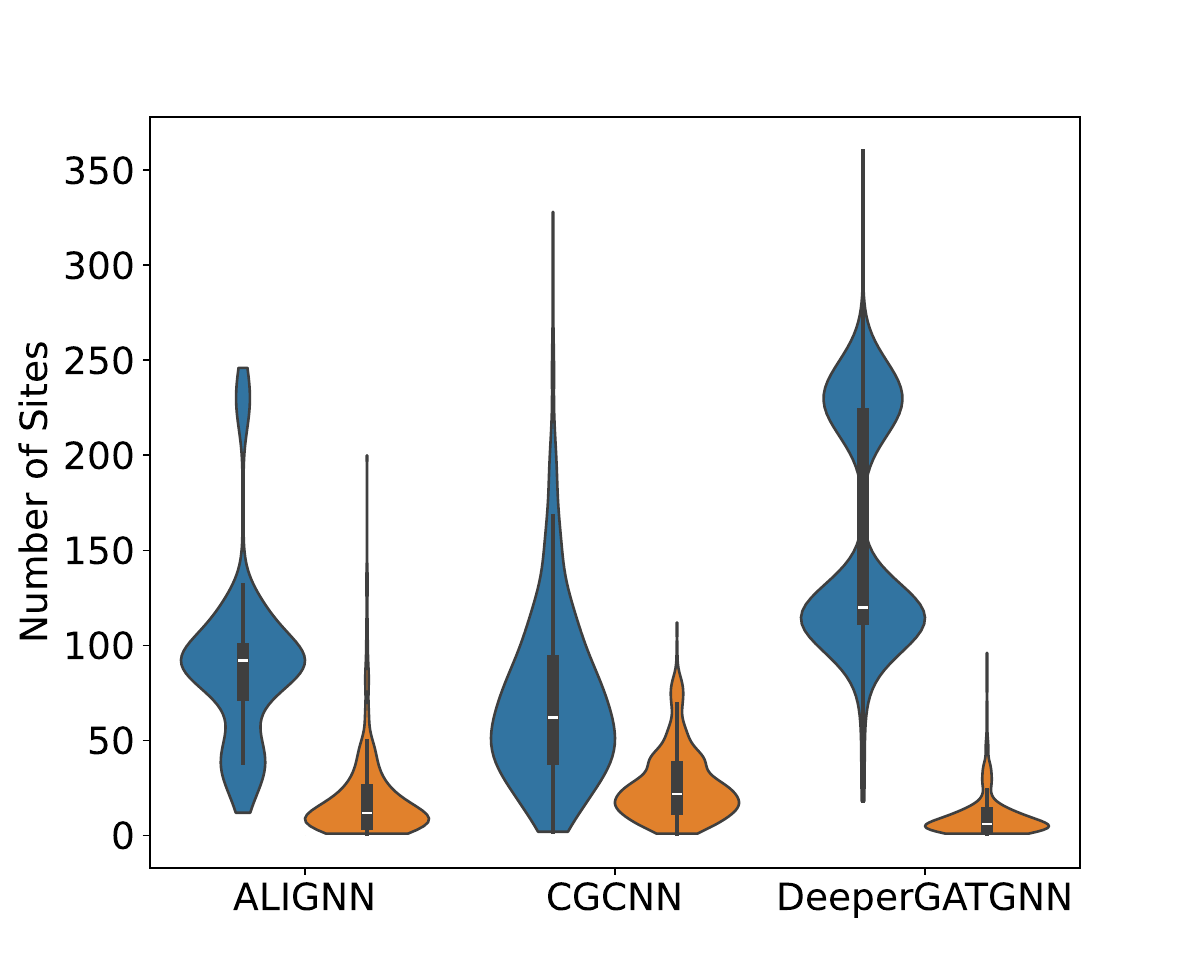}
    \end{minipage}
    \begin{minipage}[c]{0.49\linewidth}
    \includegraphics[trim={0cm 0cm 0cm 0cm},width=\textwidth]{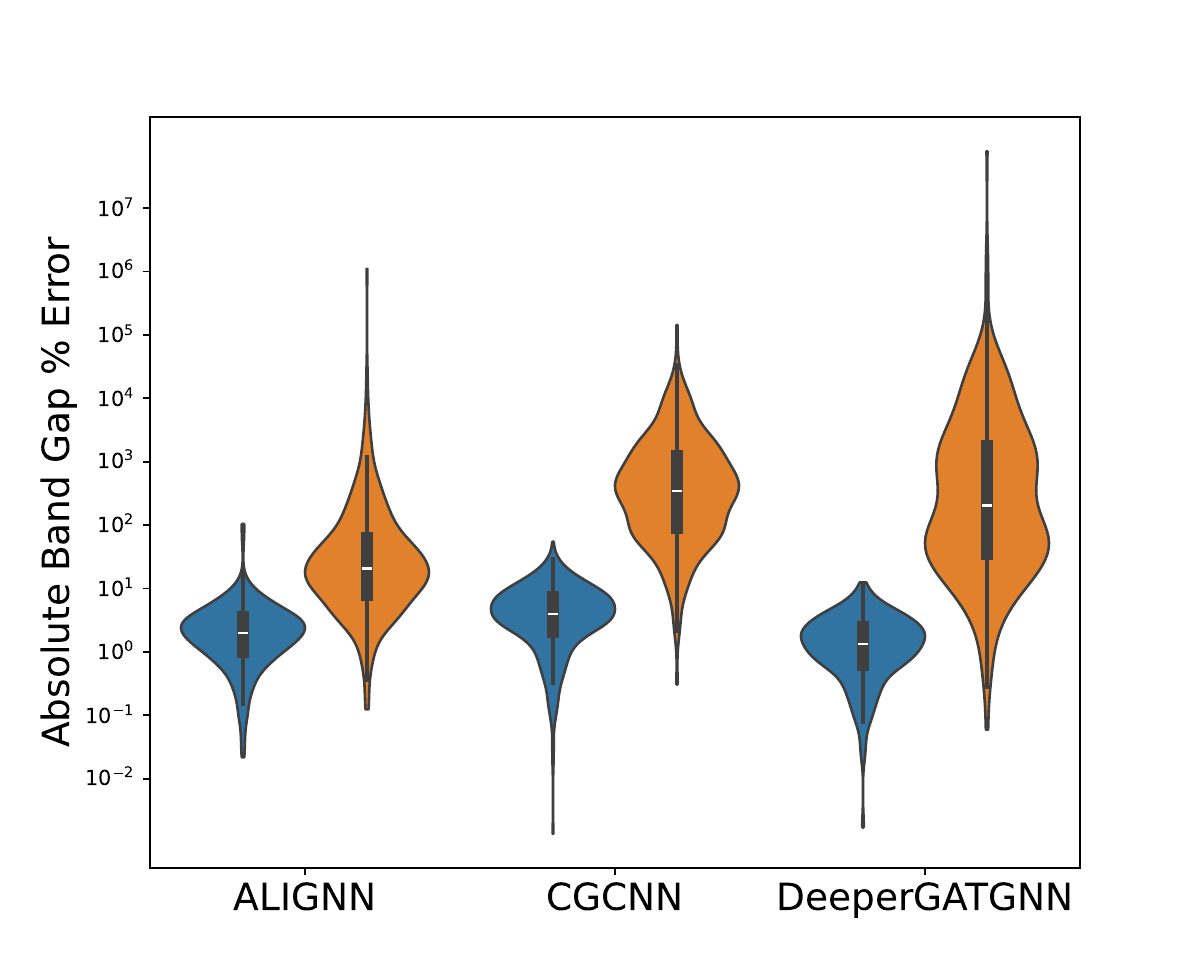}
    \end{minipage}
    \begin{minipage}[c]{0.49\linewidth}
    \includegraphics[trim={0cm 0cm 0cm 0cm},width=\textwidth]{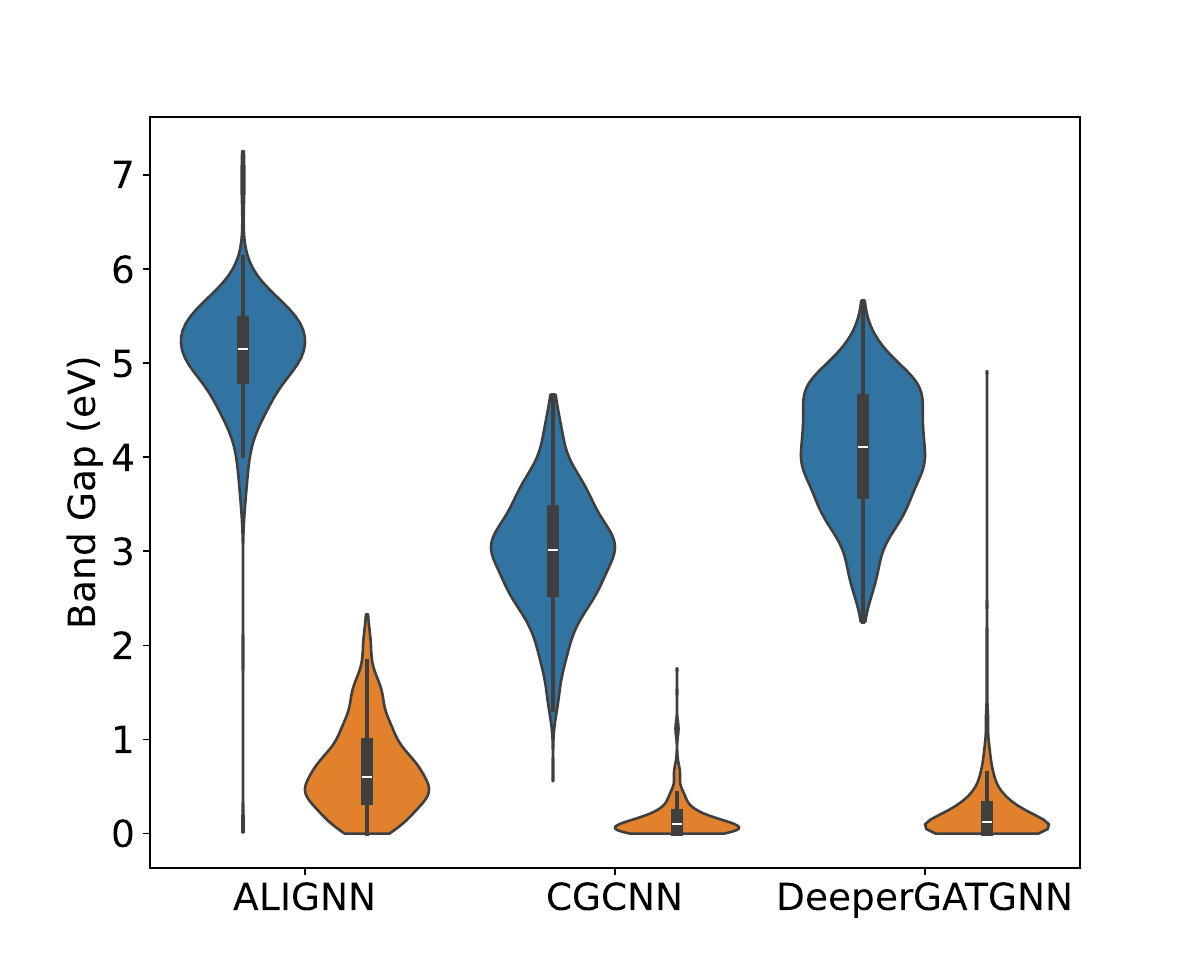}
    \end{minipage}
    \begin{minipage}[c]{0.49\linewidth}
    \includegraphics[trim={0cm 0cm 0cm 0cm},width=\textwidth]{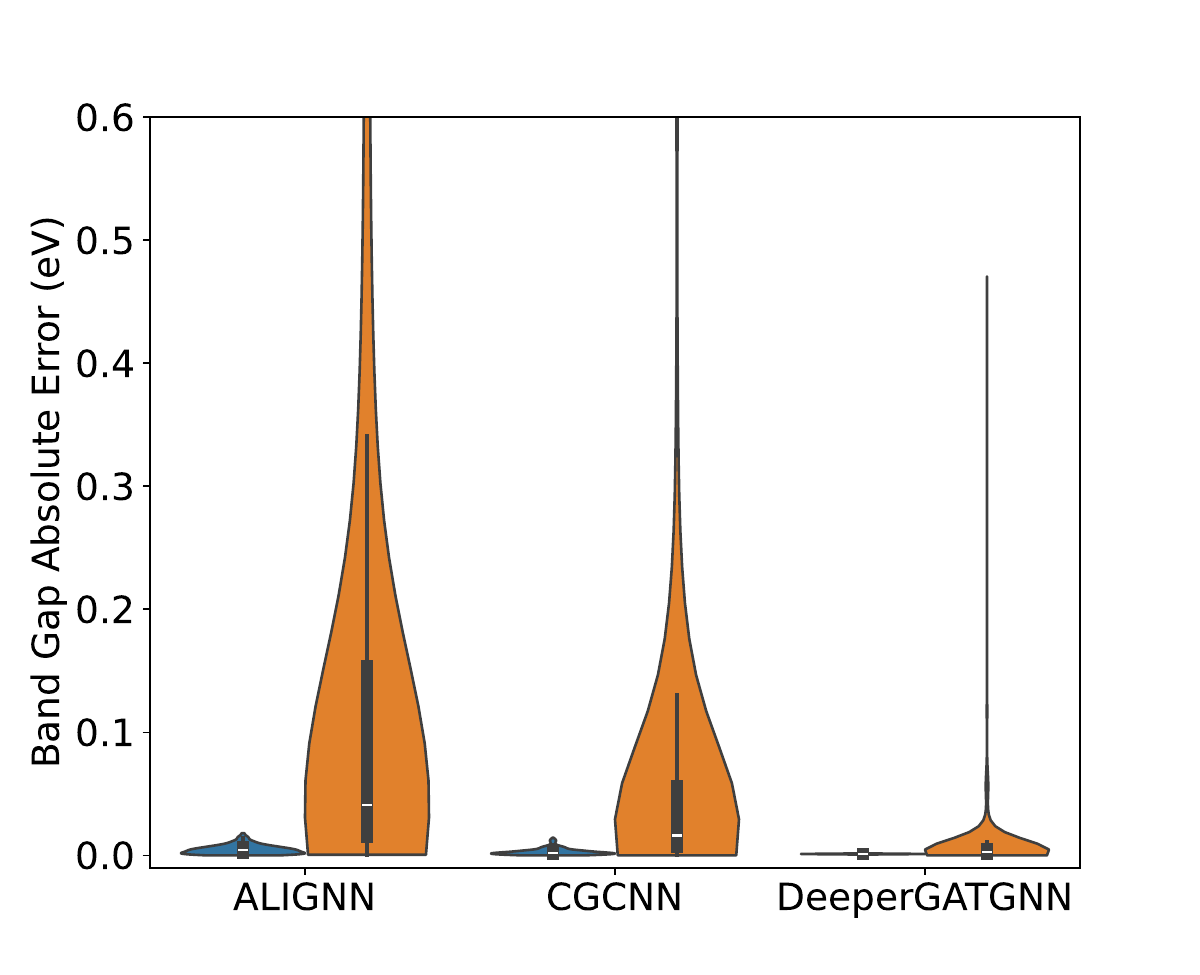}
    \end{minipage}
    \begin{minipage}[c]{0.49\linewidth}
    \includegraphics[trim={0cm 0cm 0cm 0cm},width=\textwidth]{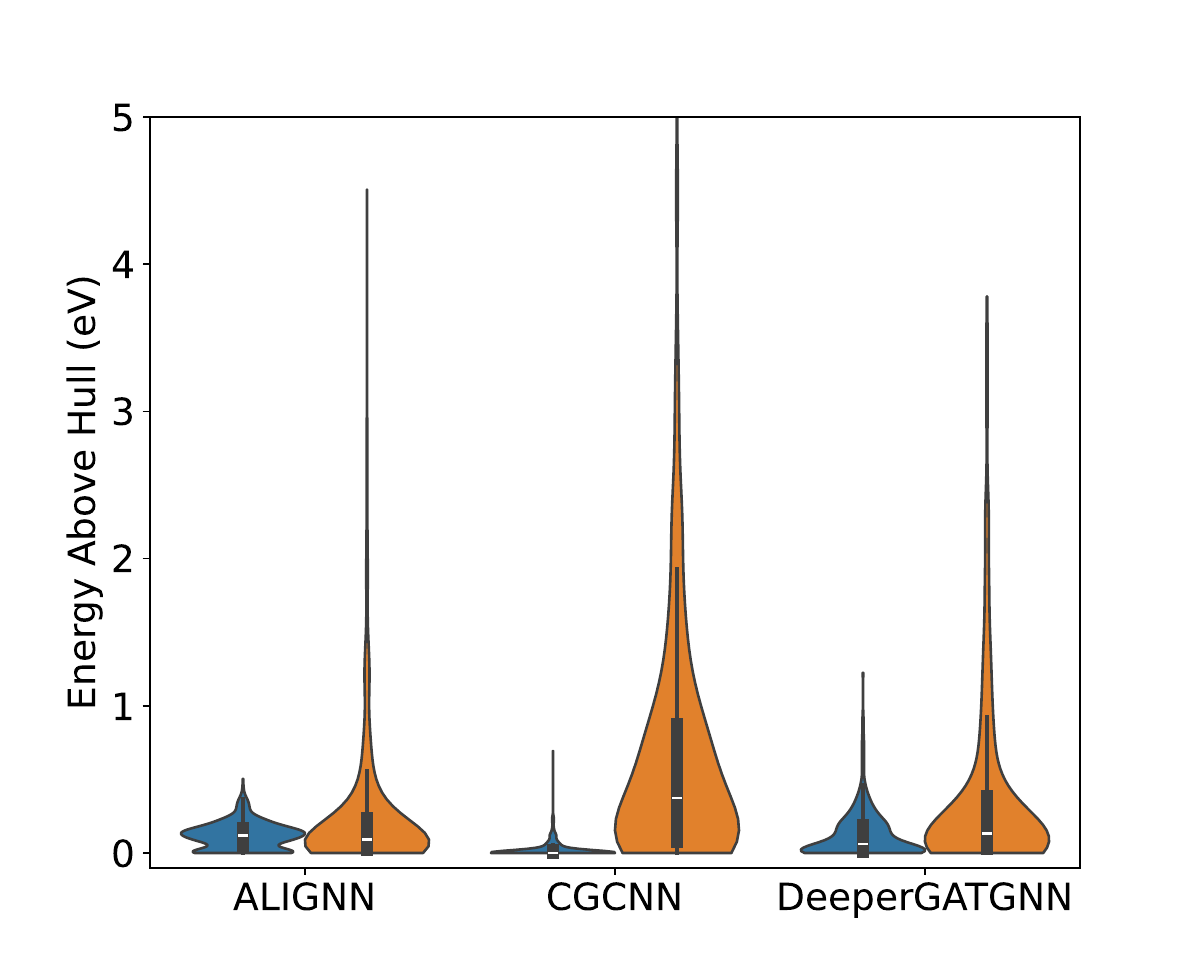}
    \end{minipage}
    \caption{Property prediction worsens on rare materials across models. \textbf{Top:} Analysis of models trained to predict formation energy, showing larger predictive errors on the $500$ rarest materials according to the Vendiscope compared to the bottom $500$ materials. The rare materials correspond to those with fewer sites in their unit cells. \textbf{Middle:} Analysis of models trained to predict band gap on non-conducting materials. Predictive errors are higher for rare materials compared to common materials. Y-axis is logarithmic. The rare materials correspond to those with smaller band gaps. \textbf{Bottom:} Analysis of models trained on band gap prediction on conductors. Prediction errors are significantly higher for rare materials. For all models except ALIGNN, rare materials correspond to those with higher energies above the hull. All distributions are statistically distinct as measured by Mann-Whitney U Tests with p-values less than $0.01$.}
    \label{fig:mat_pred_model}
\end{figure}

\parhead{Property prediction accuracy degrades on materials that enhance diversity.} The three selected models all achieved state-of-the-art property prediction performance at the time of their publication. However, they all fail to model the same types of materials: the ones that enhance diversity. 

We trained each model to predict formation energy and band gap. We then extracted embeddings from each model by using the output from the layer just before the final prediction layer and used these embeddings as materials representations for the Vendiscope. In Figure \ref{fig:mat_pred_model}, we show that the error associated with formation energy prediction is significantly higher for rare materials. The rare materials, as shown in \ref{fig:mat_pred_model}, tend to have a smaller number of sites in their unit cell. 

To characterize model behavior further, we partitioned materials into conductors (band gap $=0$ eV) and non-conductors (band gap $\ne 0$). Applying the Vendiscope to the embeddings from each group separately shows that model performance worsens significantly on rare materials. Across all models, rare materials are shown to have distinct physical properties from their bottom-scoring counterparts. One-sided Mann-Whitney U tests confirm that rare non-conductors have lower band gaps than common materials. The tests also confirm that rare conductors have large energies above the hull for both the CGCNN and DeeperGATGNN models. 

Our results confirm previous work by \cite{li2023exploiting}, who also observe strong performance on redundant materials and poorer performance elsewhere. 

\parhead{The Vendiscope detects duplicate crystals in the Materials Project database.} We also apply the Vendiscope to detect near-duplicates in the Materials Project database. We consider two embedding spaces for the Vendiscope. The first embedding space is the one implied by formation energy prediction using ALIGNN. Using Algorithm \ref{alg:duplicate}, we identify that $148,907$ materials (87.9\% of the dataset) are near-duplicates at a similarity threshold of $s=0.9$, decreasing only to $121,683$ at a stricter threshold of $s=0.95$. The second embedding space corresponds to band gap prediction using ALIGNN. Among conductors in this space, $67,910$ materials are near-duplicates at $s=0.9$, with $52,684$ remaining near-duplicates at $s=0.95$. For non-conductors, $78,643$ materials are near-duplicates at $s=0.9$, with a slight reduction to $65,891$ materials at the stricter threshold of $0.95$.

With the Vendiscope, we are able to find all of these near-duplicates rapidly: in all embedding spaces, we only need to compute $19\%$ of all pairwise similarities in the Materials Project database. Alternative approaches to identifying materials with similar structures rely on computing all pairwise similarities and require manual inputs. For example, the Materials Project database compares carefully curated coordination site fingerprints across all materials to identify crystals with similar atomic arrangements and bonding patterns. 

\section*{Analyzing State-Of-The-Art Generative Models}
\label{sec:generative-models}

Here, we apply the Vendiscope to CIFAR-10, a dataset containing $50,000$ images across 10 classes. This dataset has become a popular benchmark for training image generative models. Gaining insights into the contents of CIFAR-10 and understanding how individual images affect model performance can lead to improved modeling efforts. Using the Vendiscope, we identify the near-duplicates present in CIFAR-10 and in the outputs of generative models trained on it. We also leverage the Vendiscope to study memorization in $13$ state-of-the-art generative models spanning different generative modeling frameworks.

\begin{figure*}[t]
    \centering
    \begin{subfigure}[b]{0.56\textwidth}
        \centering
        \includegraphics[trim=0.6cm 5cm 6cm 2cm, width=\linewidth]{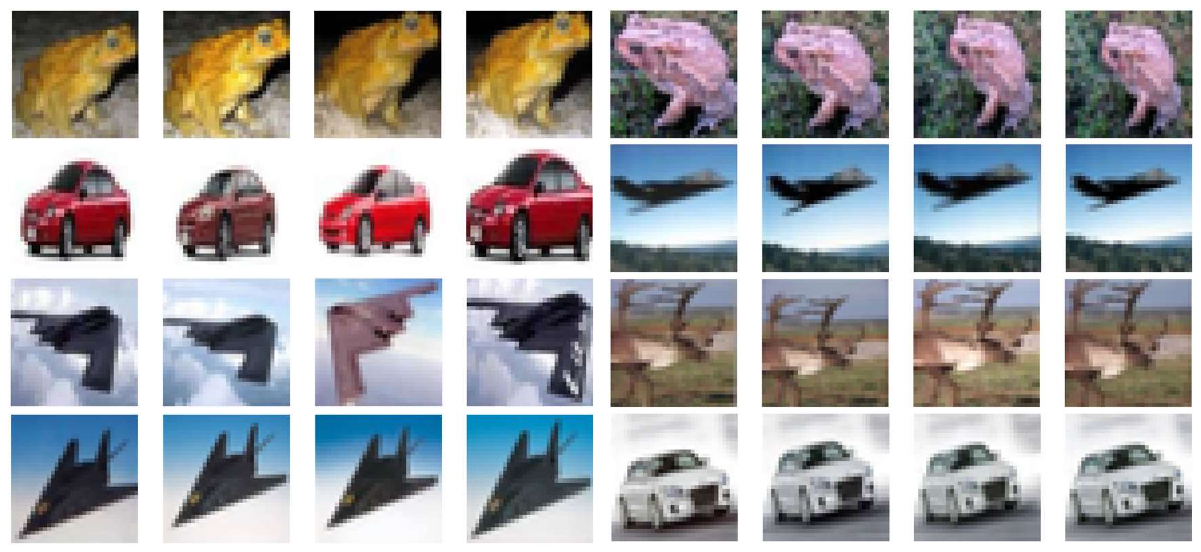}
    \end{subfigure}
    \begin{subfigure}[b]{0.43\textwidth}
        \centering
        \includegraphics[width=\linewidth]{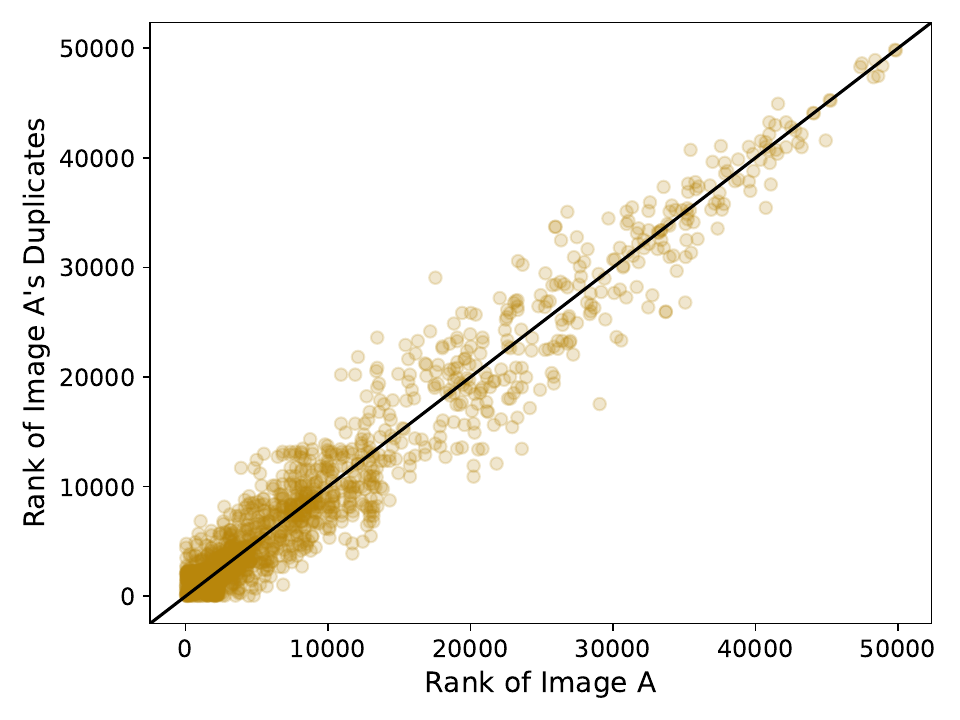}
    \end{subfigure}
    \caption{\textbf{The Vendiscope can efficiently detect near-duplicates}  \textbf{Left:} Selected near-duplicates in CIFAR-10. \textbf{Right:} The Vendiscope rank of each pair of near-duplicates is concentrated along the diagonal, demonstrating that similar images contribute similarly to a dataset's overall diversity. A total of $955$ images are near-duplicates in CIFAR-10.} 
    \label{fig:cifarTrain_Dup}
\end{figure*}

\parhead{Detecting duplicates in CIFAR-10.} CIFAR-10 is known to contain many duplicates and near-duplicates \citep{recht2018cifar}. The presence of duplicates is known to hinder model training and robustness \citep{barz2020we}. We demonstrate that the Vendiscope can identify all of these duplicates using Algorithm \ref{alg:duplicate}. We first run the Vendiscope using image embeddings from the DINOv2 ViT-L/14 network~\citep{oquab2023dinov2} and a cosine similarity kernel. 

\begin{figure}[t]
\centering
\includegraphics[width=0.8\textwidth]{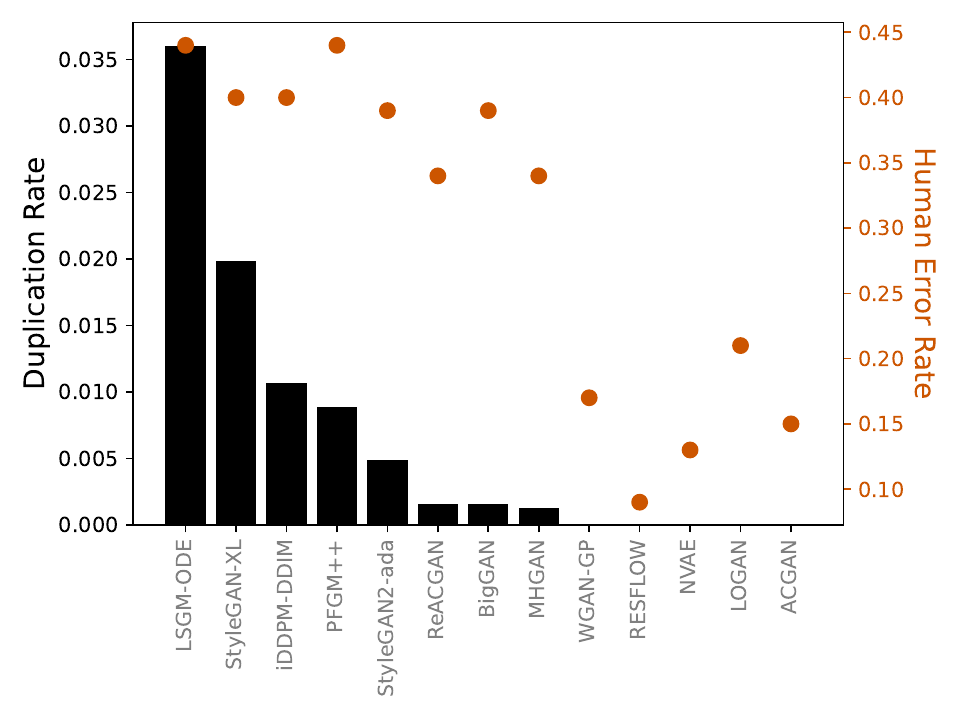} 
\caption{CIFAR-10 image generative models with high duplication rates have high human error rates. Models that produce $0$ duplicates produce lower quality outputs according to human judges.}
\label{fig:cifar10_her_dup}
\end{figure}

Once the Vendiscope weights are calculated, we apply Algorithm \ref{alg:duplicate} with a search range of $m=10,000$, which corresponds to computing only $33\%$ of all pairwise similarities. We use a similarity threshold of $s=0.95$ to identify $955$ duplicates. Fig. \ref{fig:cifarTrain_Dup} shows how images that are visually almost identical have similar contributions to the diversity of the dataset. While identifying duplicates in CIFAR-10 is feasible with naive searches or even manual annotation like in \cite{recht2018cifar}, the Vendiscope provides a scalable alternative for larger datasets where other methods are computationally impractical. 

\begin{figure}[tp!]
    \centering
    \includegraphics[trim={0cm 0 0cm 0},clip,width=\textwidth]{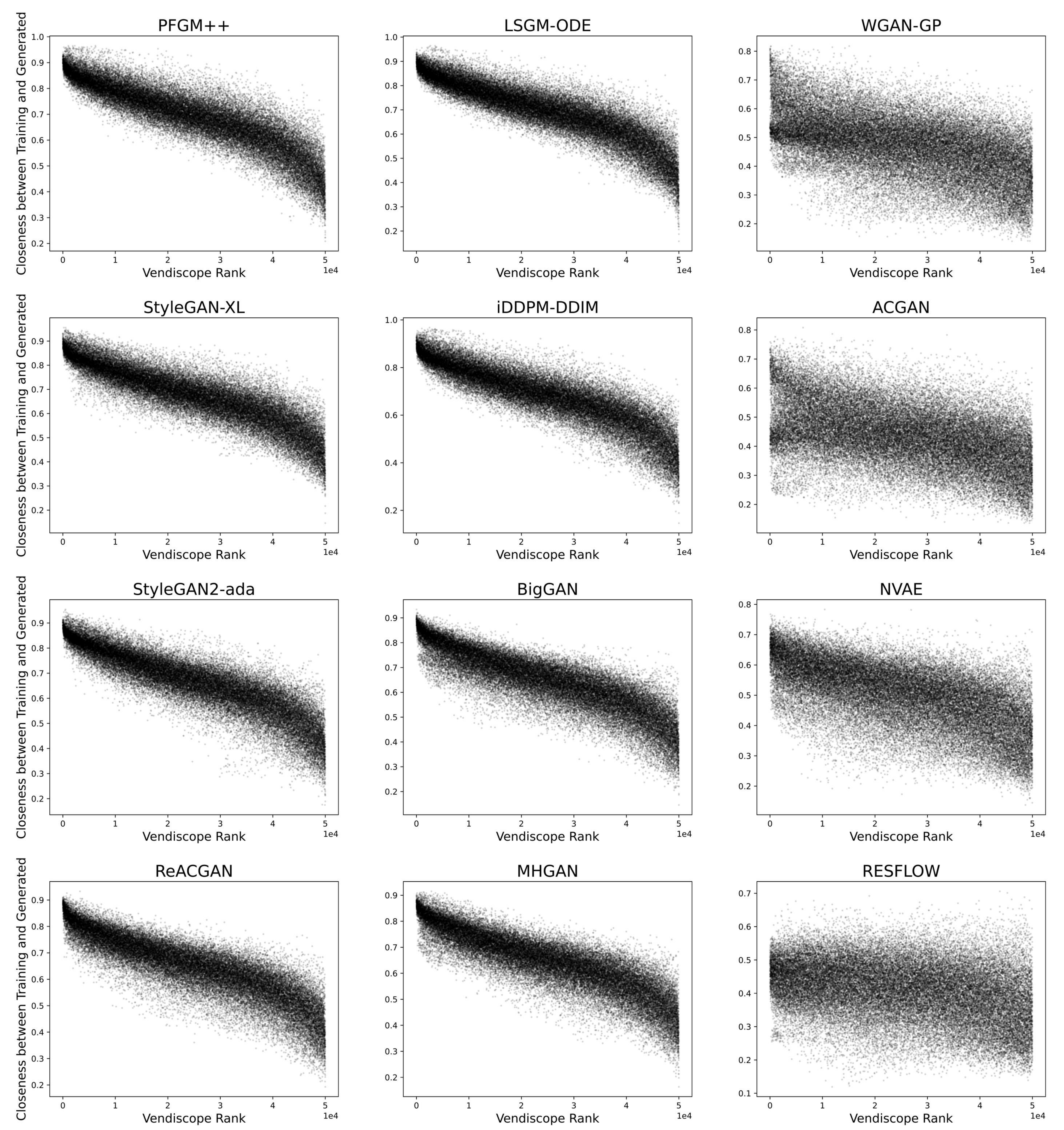}
    \begin{tikzpicture}[overlay, remember picture]
       \draw[dashed, thick, gray] ([xshift=13.1cm, yshift=-2.95cm]current page.north west) -- ([xshift=13.1cm, yshift=-18.6cm]current page.north west);
    \end{tikzpicture}
    \begin{minipage}[c]{0.98\linewidth}
    \includegraphics[trim={0.0cm 0 1.6cm 0},clip,width=\textwidth]{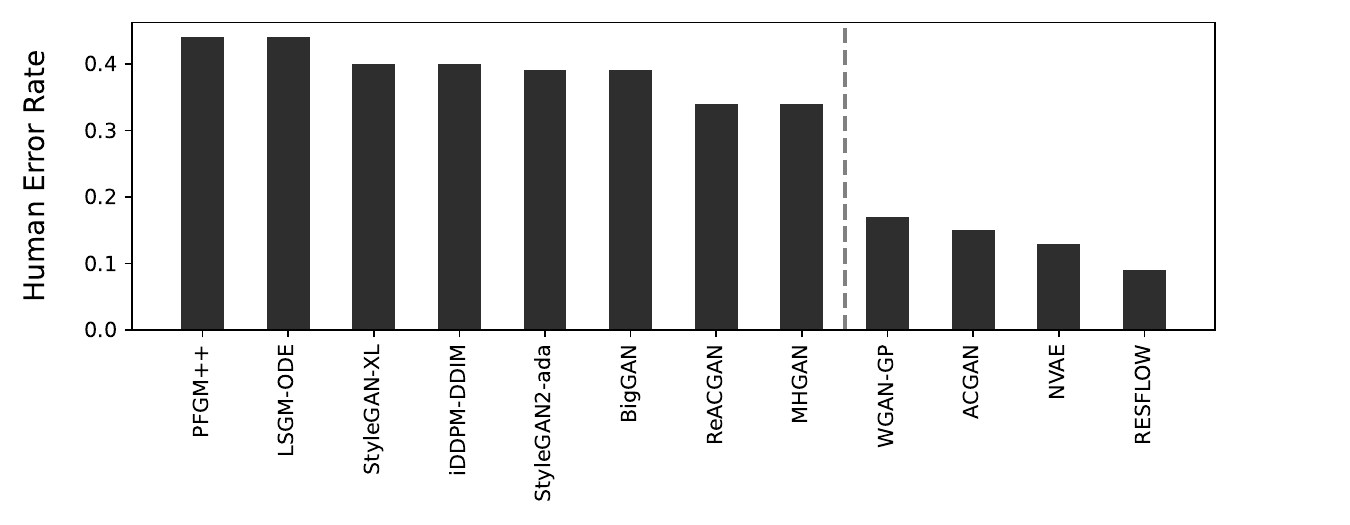}
    \end{minipage}
    \caption{Measuring memorization for different generative models trained on CIFAR-10. Here we start with the training set and look at the image they are closest to in the outputs from the generative models. The trend is clear: the images in the training set that have higher similarity to the outputs of the generative models are those that contribute least to the diversity of CIFAR-10. Memorization for an image in the training set is measured as its highest similarity to any sample in the generated outputs of a given generative model. Models for which the correlation is weaker are in the third column; they tend to produce images with lower quality, as measured by human error rate.}
    \label{fig:cifar10_all_memo}
\end{figure}
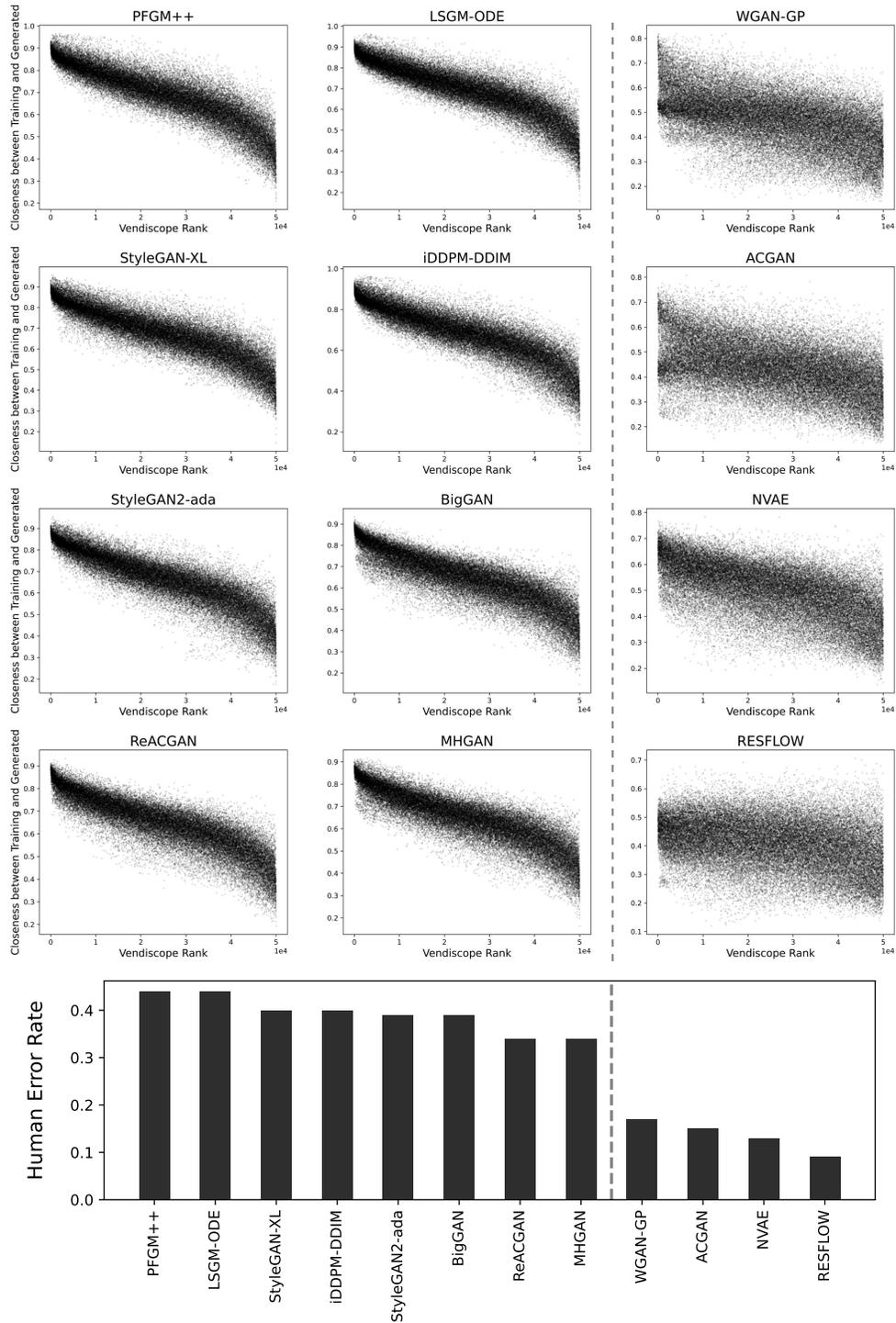

\parhead{Detecting duplicates in state-of-the-art image generative models.} We also use the Vendiscope to identify duplicates in the outputs of image generative models. \emph{Useful} generative models should produce images that are novel, diverse, and of high perceptual quality. However, existing evaluation methods for image generative models do not directly evaluate duplication in the generated outputs. Here, we evaluate the outputs from $13$ generative models trained on CIFAR-10~\citep{stein2023exposing}. Using $50,000$ samples from each model, we run the Vendiscope with cosine similarity and a Vendi Score order of $q=0.1$. We then identify duplicates using Algorithm \ref{alg:duplicate} with a search range of $m=10,000$ and a similarity threshold of $s=0.9$. In Figure \ref{fig:cifar10_her_dup}, we show the number of duplicates for each model, as well as the average human error rate provided by \cite{stein2023exposing}. We observe that the generative models producing the highest quality images, i.e. with lower human error rates, also produce a large number of duplicates.

\parhead{Detecting memorization.} Memorization is an undesirable property of generative models, although its causes are not yet well understood. We use the Vendiscope to provide insights into the types of samples that models tend to memorize. 

We compare the weight assigned by the Vendiscope to each training data point against its similarity to the generated output its most similar to. Memorized data points will have high similarity with one or more outputs from the generative model. Across all $13$ generative models, we find a strong negative correlation between the rarity of a training data point and its degree of memorization (Fig.~\ref{fig:cifar10_all_memo}). Figure~\ref{fig:memo_cifar10} shows that the samples that contribute least to the diversity of the generated outputs, i.e. those ranked lowest by the Vendiscope in terms of diversity contribution, are the ones that have highest similarity with the training set. The trend is clear: image generative models tend to memorize data points that contribute least to the diversity of the training set. This behavior has been also noted by \cite{webster2021person} in the context of image generative models for human faces. Their study found that models trained on low-diversity datasets are prone to recreate the images from over-represented identities. We additionally find that models trained on CIFAR-10 which memorize common images the most achieve higher image quality scores as measured by human error rate. (We refer the reader to Figure \ref{fig:corr_memo_cifar10} in the appendix). 

\section*{Related Work}
\label{sec:related}

The Vendiscope offers a range of capabilities, including the detection of outliers, duplicates, samples prone to memorization, and samples that models may struggle to predict---prior to any training. It also ranks data points according to their rarity. Related works only tackle a subset of these tasks, typically in specific domains, and we discuss them below.

\parhead{Near-duplicate detection.} Several methods have been developed to detect duplicates in specific domains, e.g. proteins and text~\citep{kocetkov2022stack, lee2021deduplicating, steinegger2018clustering, zhang2023retsim}. MMSeqs2 is a highly-parallelizable method for detecting duplicates in protein sequences using an approximate matching of k-mers across sequences~\citep{steinegger2018clustering}. While MMSeqs2 is fast in practice, it can be inaccurate~\citep{schutze2022nearest, ou2023recent}. We observed this inaccuracy on the UniProtKB dataset in our experiments, as discussed earlier. This limitation of MMSeqs2 is partially because proteins with matched k-mers do not necessarily describe homologs or functions~\citep{schutze2022nearest, villegas2021unsupervised}. As a result, the duplicate sequences identified by MMSeqs2 are less informative. knnProtT5 is an alternative method that relies on k-nearest neighbor search on protein embeddings~\citep{schutze2022nearest}. However, this has a higher complexity of $\mathcal{O}(N\log N)$ compared to the Vendiscope and MMSeqs2, which have linear complexity. 

A popular approach for detecting duplicates in text data is the MinHash-based Locally Sensitive Hashing (LSH) algorithm~\citep{lee2021deduplicating, kocetkov2022stack}. This method estimates the Jaccard similarity between two documents using a series of hashing functions, which makes it scalable and amenable to various streaming optimizations. However, MinHash does not account for semantic similarity and is thus sensitive to typos and other adversarial attacks on text. Another duplication detection method for text is RETSim, which trains a lightweight encoder model for near-duplicate detection~\citep{zhang2023retsim}. While the results are quite encouraging, it is not feasible for researchers to train a specialized model for each potential use-case. 

The Vendiscope can detect near duplicates at scale (see \Cref{table:DupMethods}) and applies to any domain where similarity can be defined, which makes it more general. Furthermore, the Vendiscope offers additional capabilities for analyzing data collections beyond duplication detection. 

\parhead{Detecting memorization in generative models.} Significant efforts have been made to identify the causes of memorization in generative models ~\citep{jagielski2022measuring, kandpal2022deduplicating, lee2021deduplicating, somepalli2023diffusion, tirumala2022memorization}, and this topic continues to attract considerable attention from the ML community. \citet{kandpal2022deduplicating} and \citet{lee2021deduplicating} found that the presence of duplicates can be a potential cause of model memorization. However, memorization can occur even in the absence of duplicates~\citep{somepalli2023diffusion}. Overfitting has also been thought to contribute to memorization; however, \citet{tirumala2022memorization} found examples of memorized training samples from models that had not yet overfit. Moreover, \citet{jagielski2022measuring} attributed memorization to the order in which data is presented to a model, with samples presented last being more prone to memorization. The Vendiscope provides a new framework for reasoning about memorization in ML: memorization is closely linked to the contribution of samples to diversity. Our analysis of $13$ different image generative models shows that samples contributing more to diversity are less likely to be memorized, while those prone to memorization contribute the least to diversity. Moreover, by identifying a correlation between memorization and diversity, the Vendiscope adds an extra capability: it can detect data points likely to be memorized by models even before training. This is possible because the Vendiscope can be applied to any collection, whether it's the training set or outputs from a generative model.

An additional challenge in studying model memorization is the lack of reliable metrics. Existing metrics, such as \(C_T\), calibrated \(l_2\) distance, and label memorization, often require significant tuning or can only detect specific forms of memorization \citep{meehan2020non, stein2023exposing, pondenkandath2018leveraging}. Alternatives include manually feeding prompts from the training data to the model in an attempt to identify memorized samples \citep{carlini2022quantifying}. However, selecting which prompts to feed is largely ad-hoc and relies on heuristics such as prompt length. The Vendiscope can guide such a prompting strategy by identifying clusters of data that are likely to be memorized.

\parhead{Characterizing large-scale datasets.} Certain datasets, such as The Stack, FineWeb, C4, and RedPajama, have become staples for training large language models \citep{kocetkov2022stack, penedo2024fineweb, webster2021person}. However, the contents of these datasets are not well understood, and very few works have attempted to address this gap. \citet{penedo2024fineweb} conducted extensive ablation studies to justify the curation strategies behind FineWeb, yet the dataset itself remains largely opaque beyond a high-level topic analysis and a word-association bias study. \citet{elazar2023s} analyzed numerous benchmark text datasets, revealing statistics such as the most common $n$-grams, the diversity of text sources, and the number of exact duplicate documents within each dataset. The overall pipeline enables searches for the amount of personally identifiable information (PII) present in each dataset. \citet{zhong2024explaining} also aimed to improve vision and language dataset interpretation by identifying common topics present in the dataset. The Vendiscope provides a more comprehensive analysis of the composition of a data collection and is applicable beyond documents and images.

\parhead{Vendi scoring.} 
The Vendi Score (VS) and the probability-weighted Vendi Score (pVS) were introduced by \cite{friedman2023vendi} and quantify the diversity of a collection of elements. \citet{pasarkar2023vendi} demonstrated the utility of optimizing the VS in molecular dynamics simulations, finding that jointly optimizing the VS and the energy of a molecule significantly speeds up and diversifies the exploration of complex energy landscapes. \cite{berns2023towards} optimized the sum of the pVS and the Shannon entropy of the probabilities involved in the computation of the pVS to balance the modes of generative models, enhancing their ability to produce diverse outputs in artistic domains. The VS has been extended and applied in multiple ways, owing to its flexibility~\citep{askari2024improving, kannen2024beyond,liu2024diversity,nguyen2024quality, mousavi2024vsi, pasarkar2024cousins, rezaei2025alpha, bhardwaj2025robust}. The Vendiscope maximizes the pVS and enables scalable detection of outliers, duplicates, samples prone to memorization, and samples that models may struggle to predict, even before training. It also ranks the data points in a given dataset according to their rarity and provides per-data-point weights that can aid in data processing, model training, and evaluation.

\begin{table}[]
\def\arraystretch{1.1}
\centering
\begin{NiceTabular}{@{}m{0.22\linewidth}m{0.24\linewidth}m{0.22\linewidth}m{0.17\linewidth}@{}}[colortbl-like]
\multirow[b]{2}{*}{\textbf{Method}} & \multirow[b]{2}{*}{\textbf{Input}}  & \multicolumn{2}{l}{\textbf{\ \ \ \ \ \ \ \ \ \ \ \ Complexity}} \\ \cline{3-4}
                                   &                         & \multicolumn{1}{l}{\textbf{Time}} & \multicolumn{1}{l}{\textbf{Space}}  \\ \hline
\rowcolor{gray!50}
MMSeqs2                 & Protein Sequences                        & O(N)  &  O(NL)\\ 
knnProtT5 & Protein Embeddings                        & O(N$\log$N) & O(ND) \\                                  
\rowcolor{gray!50}
MinHash                            & Raw Text & 
O(KT$^2$N)                       & O(NK) \\ 
RETSim                  & Text Embeddings                        &  O(ND)    &  O(ND)                                       \\
\rowcolor{gray!50}
The Vendiscope                  &  Any Embedding                      &  O(Nm+ND$^2$)    & O(ND)               
\end{NiceTabular}
\caption{A comparison of the complexity of various de-duplication methods for a dataset with $N$ samples. For protein sequence databases, we denote by $L$ the maximum protein sequence length. For embedding-based methods, we denote by $D$ the dimensionality of each sample's embedding. For MinHash, we denote by $K$ the number of hashing functions used, and by $T$ the maximum number of tokens in a document. For the Vendiscope, we denote by $m$ the search-range used in Algorithm \ref{alg:duplicate}. The Vendiscope has linear time and space complexity and is more general.}
\label{table:DupMethods}
\end{table}

\section*{Discussion And Conclusion}
\label{sec:discussion}

In this work, we introduced the Vendiscope, a novel computational tool designed to enhance our ability to analyze large, complex datasets. By leveraging the probability-weighted Vendi scores, the Vendiscope enables researchers and practitioners to systematically measure and weigh the contribution of individual data points to the overall diversity of a dataset. This simple capability in turn can serve to detect outliers, duplicates, samples prone to memorization, and samples that models may struggle to predict. This has profound implications for discovery and AI.


Our results reveal several key insights. First, we showed that, in the context of the protein universe, a vast majority of sequences—over $200$ million out of $250$ million—are near-duplicates, underscoring the redundancy inherent in biological data. Importantly, we demonstrated that existing ML models, such as AlphaFold, struggle to predict proteins associated with Gene Ontology (GO) functions that contribute most significantly to the diversity of the dataset. This provides insights into the limitations of current methods and emphasizes the need for more nuanced data collection, model development, and evaluation in computational biology.

Similarly, our application of the Vendiscope to the Materials Project database revealed that over $85\%$ of crystals with formation energy data are near-duplicates. Moreover, we found, similarly to the finding we had with the protein universe, that ML models often perform poorly on the materials that enhance diversity, an insight that could guide future materials discovery and design processes. These findings emphasize the importance of considering diversity in both data collection, model development, and evaluation.

One of the most compelling aspects of the Vendiscope is its ability to characterize data points that are most prone to memorization in generative models—a pressing concern in AI. Our results show that the best-performing generative models often memorize training samples that contribute the least to the overall diversity of the dataset, revealing a new aspect of model behavior. By identifying data points that are prone to memorization even before training takes place, the Vendiscope presents a powerful means of improving AI model development.

Beyond its applications in individual domains, the Vendiscope provides a unified framework for analyzing complex data at scale. It helps identify duplicates, outliers, poorly represented data points, and samples that may challenge models—even before training begins. These capabilities offer significant benefits for data preprocessing, model development, and evaluation. Researchers, engineers, and data auditors can use the Vendiscope to audit datasets, identify potential biases, and refine data collection practices. For AI ethicists, the Vendiscope offers a critical lens to understand how models interact with data, particularly in the context of bias, memorization, and data fairness, enabling better mitigation strategies to prevent undesirable outcomes in AI deployment. For scientists, the Vendiscope represents a new companion in the discovery process. 

The Vendiscope has the potential to reshape how we approach data preprocessing, model training, and evaluation. Moreover, the tool’s ability to scale---working with datasets as large as 250 million protein sequences---makes it an invaluable asset for a wide range of researchers and engineers across disciplines. By offering a way to systematically analyze data at scale, it helps unlock new possibilities for scientific discovery, model development, and ethical AI.

\section*{Methods}
\label{sec:details}

The Vendiscope enables scalable analysis of large data collections across disciplines by quantifying the contribution of each element to the overall diversity of the collection. Below, we describe the algorithms and implementation details that drive its efficiency.

\begin{algorithm}[t]
\DontPrintSemicolon
 Inputs: Data $\left\{\bx_1, \dots, \bx_n\right\}$, similarity kernel $k$, order $q > 0$, step sizes $\epsilon_1, \dots, \epsilon_n$\;
 Form a data matrix $\bX \in \mathbb{R}^{n\times d}$ and normalize its rows: $\bX_i = \bx_i/||\bx_i||_2$\;
Initialize diversity contribution scores uniformly $p_i=\frac{1}{n}$ for all  $i=1,\dots,n$\;
  \While{not converged}{
      \If{k is cosine similarity}{
      Compute weighted similarity matrix 
      $\tilde{K} = \bX^\top \text{diag}(\sqrt{p})\text{diag}(\sqrt{p})\bX$\; 
      }
      \Else{
      $\tilde{K} = \text{diag}(\sqrt{p})\bK\text{diag}(\sqrt{p})$\; 
      }
      Compute loss function $\mathcal{L}(p) = -\log \text{pVS}_k(\bx_1, \dots, \bx_n)$\;
      Compute gradients $\nabla_{p_1}\mathcal{L}(p), \dots, \nabla_{p_n}\mathcal{L}(p)$ using backpropagation\;
      Compute unnormalized weights $y_1, \dots, y_n$ such that $y_i = p_i - \epsilon_i \nabla_{p_i}\mathcal{L}(p)$\;
      Set $v_i = y_i$ for all $i$ and $\rho = \frac{1}{n}\sum_{i=1}^{n} y_i - 1$\;
      \While{the norm of v continues to change}{
      Set $v_i = \mathbb{I}(y_i > \rho)$ and $\rho = \frac{\sum_{i = 1}^{n} v_i - 1}{\sum_{j=1}^{n} v_j}$ for all $i\in \left\{1, \dots, n\right\}$\;
      }
      Update diversity contribution scores 
      $p_i = \text{max}(y_i - \rho, 0)$ for all $i\in \left\{1, \dots, n\right\}$\;
  }
 \caption{The Vendiscope: An algorithmic microscope for data collections}\label{alg:vendiscope}
\end{algorithm}

\parhead{Scalability in time and space.} 
Each iteration of the Vendiscope requires calculating the pVS for a collection of $n$ elements, which involves computing the eigenvalues of an $n \times n$ matrix. This process has a time complexity of $O(n^3)$. However, \cite{friedman2023vendi} indicated that when data embeddings are available the VS, and also the pVS, can be computed cheaply by using a cosine similarity kernel with corresponding similarity matrix $\bK = \bX^T \bX$, where $\bX \in \mathbb{R}^{n \times d}$ denotes the data embedding matrix. In this case the complexity is reduced to $O(d^2 n)$. The improvement in complexity enables the scaling of the Vendiscope to large collections where $n \gg d$. 

Once the pVS and its gradients are computed, we perform projected gradient descent. We use the active set method described in \cite{michelot1986finite}, which has an observed runtime of $O(n)$ \citep{condat2016fast}. There are certain settings for which the runtime can be quadratic  \citep{cominetti2014newton}, but we will not encounter such instances when initializing most weights to be similar. Empirically, sufficiently small learning rates ensure linear runtimes. We reach a time complexity of $O(d^3 + d^2 n + n )=O(d^2 n)$ and a space complexity of $O(dn)$ for each iteration of the Vendiscope. 

\parhead{Scaling to very large datasets with parallel computing.} As presented, Algorithm \ref{alg:vendiscope} would require the entire dataset to be loaded into memory. This is prohibitively expensive for many of the massive datasets available today. We circumvent this problem by estimating the pVS using only a subset of the data's dimensions at each iteration. In particular, at a given iteration $t$, we sample a random subset of the columns of the data matrix, $d_t \subseteq \left\{1, \dots, d\right\}$ and use $X_{d_t} \in \mathbb{R}^{n \times |d_t|}$ instead of the entire dataset $X$. This approach provides an approximation of the true pVS. 

Our subsampling approach also allows us to take advantage of parallel computing to speed up the Vendiscope. Indeed, we can use data parallelism across multiple GPUs to iterate through all dimensions faster. Each GPU maintains a copy of the Vendiscope weights and is given a separate set of data dimensions for which the GPU computes local gradients. Then, the gradients are averaged across GPUs, creating global gradients that are communicated back to each GPU. All the GPUs will therefore have the same gradients, and we can update the Vendiscope weights across each GPU. 

Algorithm \ref{alg:duplicate} is also amenable to computing optimizations. After computing the Vendiscope weights, we can divide the dataset across independent processes and run Algorithm \ref{alg:duplicate} on each subset separately. Because we only compare a given sample against a small subset of the entire dataset, we are not required to ever load the entire dataset into memory at once. Finally, we can employ batch processing and process multiple data points simultaneously by taking advantage of fast GPU matrix operations. 

\begin{algorithm}[t]
\DontPrintSemicolon
Inputs: Data $\left\{\bx_1, \dots, \bx_n\right\}$ sorted in order of the Vendiscope scores, similarity kernel $k$, near-duplicate similarity threshold $s \leq  1$, and search-range $m \leq n$\;
\For{$i=1,\dots,n$}{
      If $\bx_i$ is in a cluster $c\in \bC$ then $\bx_i$ is already analyzed, go to the next sample\;
      Else create new cluster $c \gets \left\{\mathbf{x}_i\right\}$\;
        \For{$j=i+1,\dots,i+m$}{
            If $k(\bx_i, \bx_j)>s$ and $\bx_j$ not in a cluster then 
                $c = c \cup \bx_j$\;
        }
        Add $c$ to a list of clusters $\bC$\;
}
\caption{Efficient near-duplicate detection with the Vendiscope}\label{alg:duplicate}
\end{algorithm}

\parhead{Initialization and identifiability.} In Algorithm \ref{alg:vendiscope}, we initialize the weights to be equal, reflecting, in a Bayesian sense, an uninformative prior over the Vendiscope's probabilities. This choice allows the Vendiscope to assign identical weights to exact duplicates. A random weight initialization would cause issues with the identifiability of exact duplicates due to the optimization of the pVS. For exact duplicates, an optimized pVS only places a constraint on the sum of their scores. Consider, for instance, a collection with 3 elements and the kernel matrix with the first column $\bx_1 = [1 \text{ } 0 \text{ }0]^T$ and identical second and third columns $\bx_2 = \bx_3 = [0 \text{ } 1 \text{ } 1]^T$. In this setting, the pVS can be maximized with $p_1=0.5$ and $p_2+p_3=0.5$, yielding an optimal pVS of $2$. If we initialize the weights to be equal, $p_2$ and $p_3$ will have identical gradients throughout the iterations and will remain equal.

\parhead{Hyperparameter selection and convergence criteria.} The Vendiscope requires specifying the order $q > 0$ of the pVS. Previous work by \cite{pasarkar2023vendi} demonstrated that small values of $q$, are more sensitive to rare features in the collection, whereas large values of $q$ place greater emphasis on more common features. We find that choosing small values of $q$ helps avoid the sparsity issue that may arise in large collections, where most data points have weight zero due to limitations in numerical resolution. We use $q=0.1$ and $q=0.5$ in our experiments.

We use a convergence criterion based on the learned ranking of the data points in the collection and stop the Vendiscope when the ranking stabilizes. In all of our studies, we found this typically occurs within $500$ iterations.

\section*{Code and Data Availability}
\label{sec:code-availability}
We provide a smaller version of the UniProtKB database, curated using the Vendiscope for de-duplication. Additionally, we publish the learned Vendiscope weights for each protein in UniProtKB. Both the data and the weights can be downloaded from the \href{https://drive.google.com/drive/u/9/folders/1zQdK7CfUBHt8wsQDWV57Q-paKKA37VfP}{Vertaix Drive}. 

All code and model checkpoints will be made available upon publication.

\subsection*{Acknowledgements}
We thank the members of Vertaix for their comments. We thank Cindy Zhang for her input at the very beginning of this project. Amey Pasarkar is funded by the NSF GRFP fellowship. Adji Bousso Dieng is supported by the NSF, Office of Advanced Cyberinfrastructure (OAC): \#2118201. She also acknowledges Schmidt Sciences for their AI2050 Early Career Fellowship.

\bibliographystyle{apa}
\bibliography{arxiv}

\section{Appendix}
\label{app:proofs}

\subsection{Material Property Prediction Model Training}
We train 3 models on the Materials Project (v2024.12.18). We use the recommended settings from each model for pre-processing crystal structures. We therefore use a cutoff radius of $8$ \r{A} for constructing graphs for CGCNN and DeeperGATGNN, and $4$ \r{A} for constructing graphs for ALIGNN. We sweep over hyperparameters such as the number of hidden layers and hidden dimensions before training models on the entire dataset. All models are trained to convergence: CGCNN uses $1000$ epochs with batch-size $256$, DeeperGATGNN uses a batch-size of $100$ for $400$ epochs, and ALIGNN uses a batch-size of $16$ for $300$ epochs. We use the model checkpoint at the final epoch for all downstream analysis. 

\subsection{Additional Image Generative Model Analysis}

We studied popular model architectures, including diffusion models, GANs, VAEs, and flows. In all, we tested 8 GAN models: ACGAN \citep{odena2017conditional}, BigGAN \citep{brock2018large}, LOGAN \citep{wu2019logan}, ReACGAN \citep{kang2021rebooting}, MHGAN, \citep{turner2019metropolis}, WGAN-GP \citep{gulrajani2017improved}, StyleGAN2-ada \citep{karras2020training}, and StyleGAN2-XL \citep{sauer2022stylegan}. Additional models tested include NVAE \citep{vahdat2020nvae}, RESFLOW \citep{chen2019residual}, and the three diffusion models iDDPM-DDIM \citep{nichol2021improved} PFGM++ \citep{xu2023pfgm++}, and LSGM-ODE \citep{vahdat2021score}.

\begin{figure*}[t]
\centering
\includegraphics[width=0.9\textwidth]{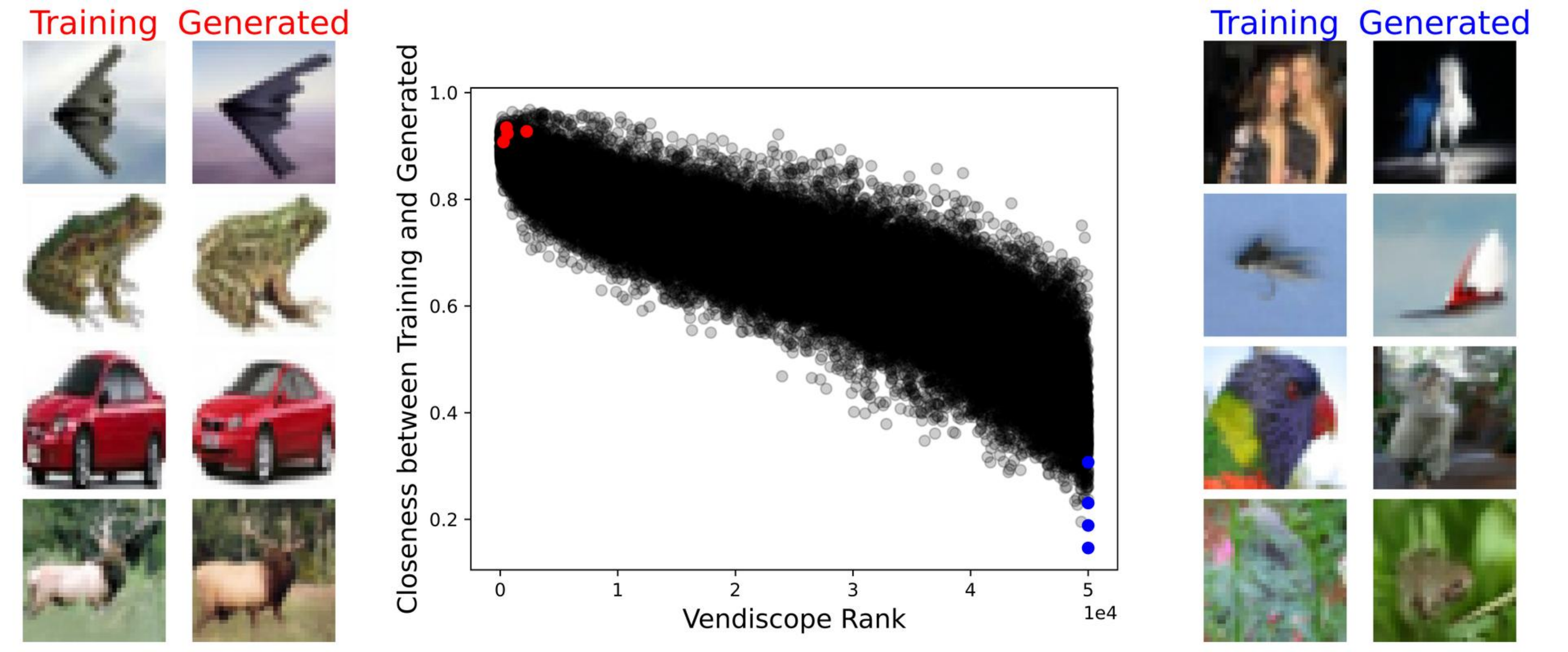}
\caption{Analyzing memorization of iDDPM-DDIM, a diffusion model, trained on CIFAR-10. The Vendiscope weights are strongly correlated with the similarity of the generated outputs to the training set. \textbf{Left:} Examples of redundant training data points, those identified by the Vendiscope to contribute least to diversity, are memorized by the generative model. The red dots in the middle figure show these samples. \textbf{Right:} Rare samples, those with high contributions to diversity, are not memorized, the training data points closest to them look visually very different. The blue dots in the middle figure show these samples, they have the lowest similarity to the training set. } 
\label{fig:memo_cifar10}
\end{figure*}

In Figure \ref{fig:memo_cifar10}, we show some examples of the varying degrees of memorization for rare and common samples on the iDDPM-DDIM model from \cite{nichol2021improved}. The rare samples in the training dataset are not represented in the generated dataset, whereas the model generates almost exact replicas of common samples. 

\begin{figure*}[t]
\centering
\includegraphics[trim={0cm 0 0cm 0},clip,width=\textwidth]{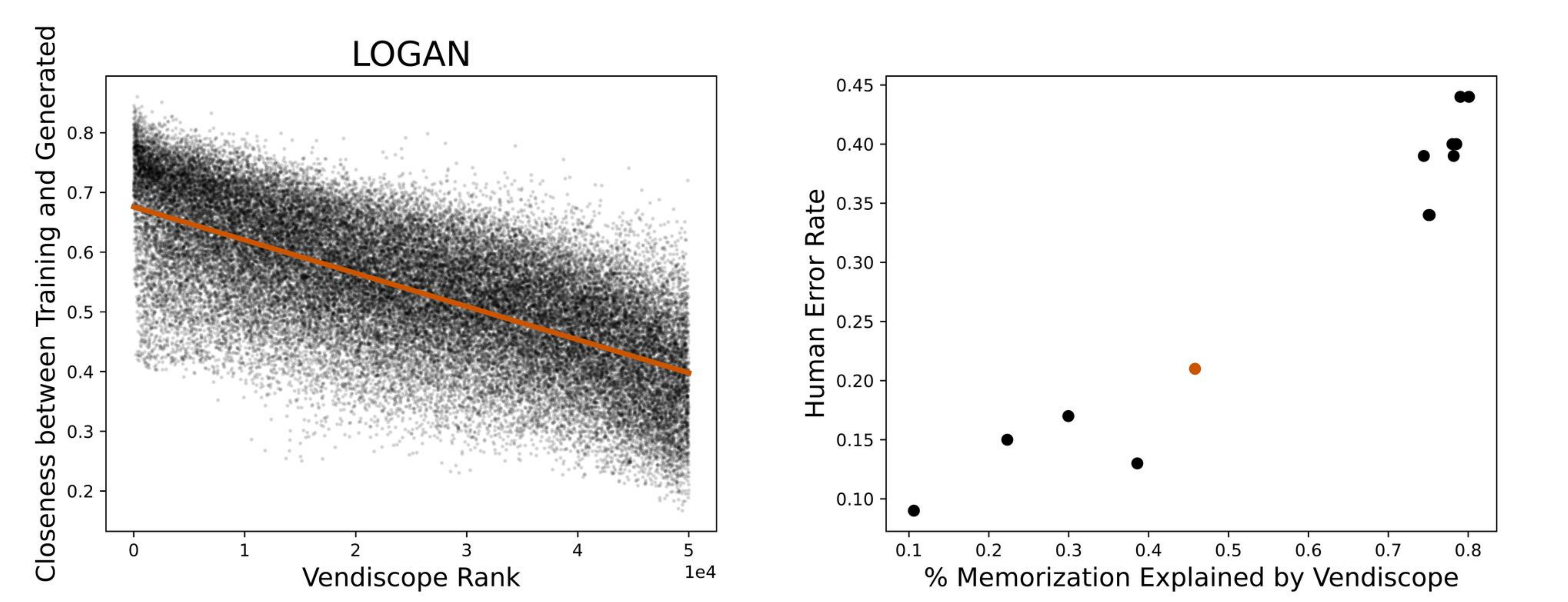}
\caption{Image fidelity is linked to the memorization of the most common samples. \textbf{Left:} A scatter plot showing the ranking learned by the Vendiscope on the CIFAR-10 training data against their degree of memorization by LOGAN. The red line corresponds to a linear regression fit and shows a negative correlation between the two. \textbf{Right:} A scatter plot of the Human Error Rate for all $13$ models (LOGAN highlighted in orange) against the degree of memorization explained by the Vendiscope's ranking of the training data. \% Memorization explained by the Vendiscope is measured by computing the $R^2$ between the Vendiscope's ranking of CIFAR-10 training data and the closeness to the nearest generated output.} 
\label{fig:corr_memo_cifar10}
\end{figure*}

\begin{figure}[!t]
    \centering
    \includegraphics[trim={0cm 0 0cm 0},clip,width=\textwidth]{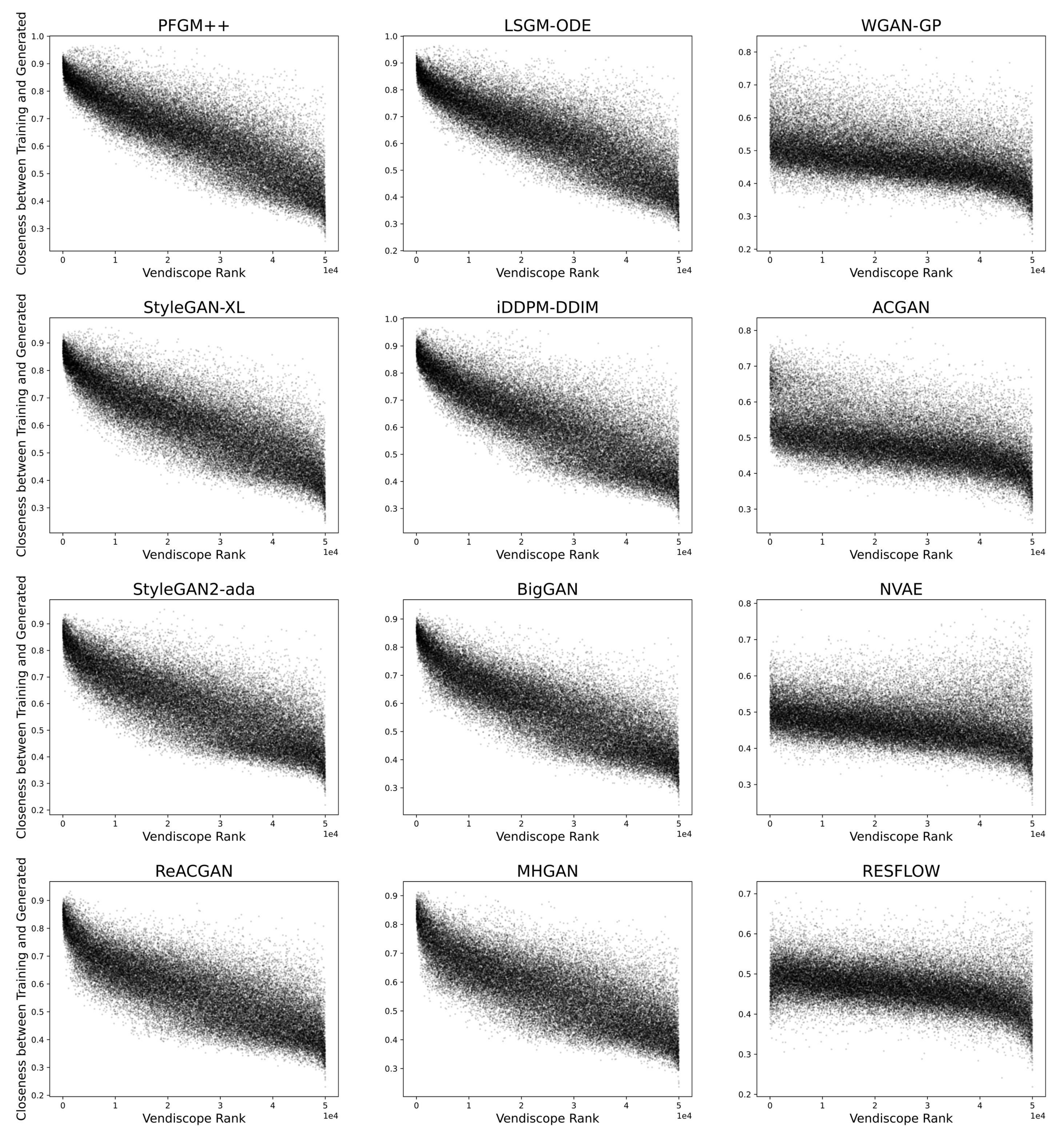}
    \begin{tikzpicture}[overlay, remember picture]
        \draw[dashed, thick, gray] ([xshift=13.1cm, yshift=-6.1cm]current page.north west) -- ([xshift=13.1cm, yshift=-19.2cm]current page.north west);
    \end{tikzpicture}
    \caption{Measuring memorization by looking at the generated collection from different generative models trained on CIFAR-10. Here we start with the generated collection instead of the training set as we did in the main draft. The conclusion is the same: generated images that have higher similarity to the training set are those that contribute least to the diversity of the generated collection. Memorization for a generated image is measured as its highest similarity to any sample in the training set. Models for which the correlation is weaker are in the third column. They correspond to models that tend to produce images with lower perceptual quality, as measured by human error rate.}
    \label{fig:cifar10_model_memo}
\end{figure}
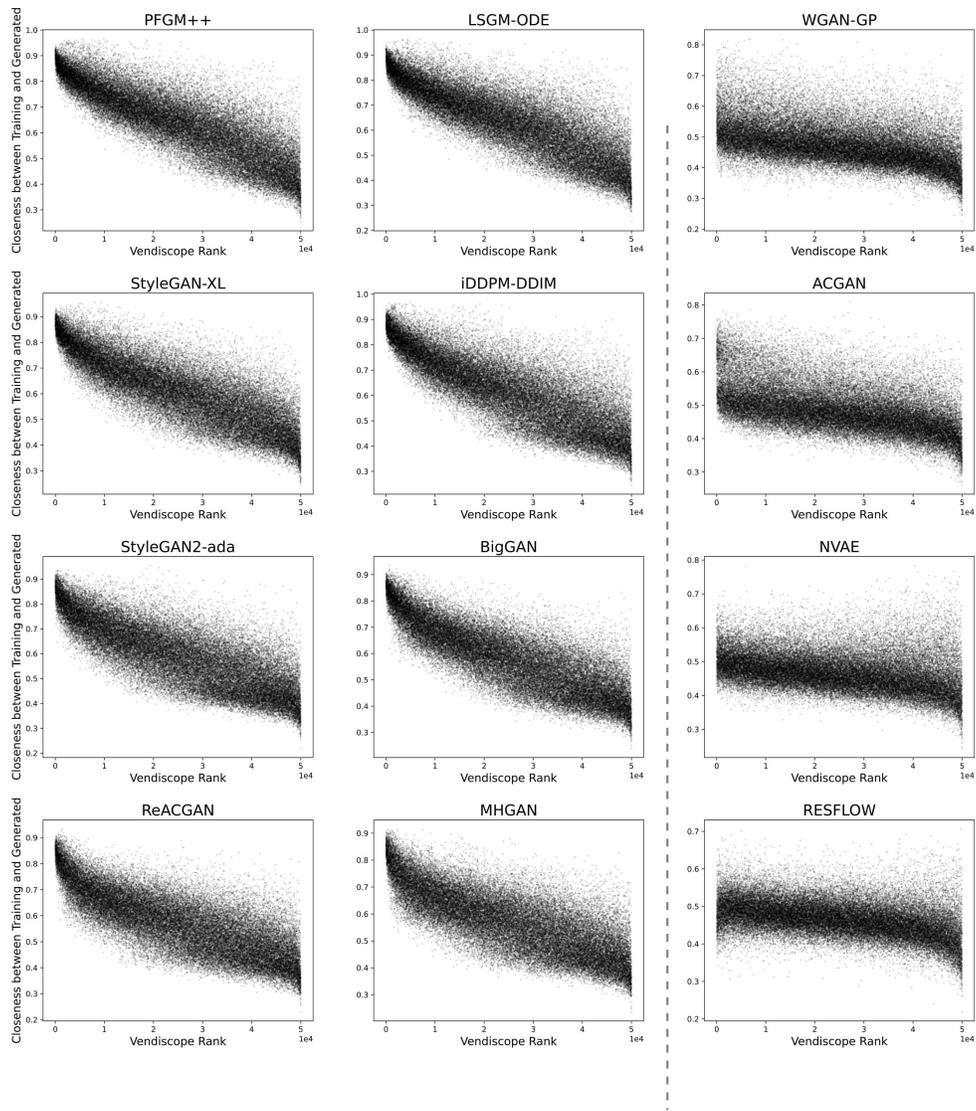

We also find that models that memorize training data are those that generate the highest-quality images (Figure \ref{fig:corr_memo_cifar10}). These models memorize common images. Models such as LOGAN that do not memorize, do so at the cost of producing images of lower quality as measured by human error rate. Finally, in Figure \ref{fig:cifar10_model_memo}, we run the Vendiscope on the generated outputs from each model. We find that the most common images generated by each model are those that have high similarity with the training data. 

\end{document}